\pdfoutput=1

\documentclass[11pt,table,xcdraw]{article}
\makeatletter
\def\input@path{{latex/}{latex/tikz/}{latex/algorithms/}{latex/tables/}}
\makeatother

\usepackage[preprint]{latex/acl}

\usepackage{times}
\usepackage{latexsym}

\usepackage[T1]{fontenc}

\usepackage[utf8]{inputenc}

\usepackage{microtype}

\usepackage{inconsolata}

\usepackage{graphicx}
\usepackage{amsmath}
\usepackage{amssymb}
\usepackage{amsthm}
\usepackage{enumerate}
\usepackage{float}
\usepackage{graphicx}
\usepackage{algorithm}
\usepackage[indLines]{algpseudocodex}
\usepackage{todonotes}

\graphicspath{{latex/images/}}
\usepackage{tikz}
\usetikzlibrary{calc}
\usepackage{xcolor}

\newcommand{\Strings}{\mathcal{S}}
\newcommand{\Seq}{\mathrm{Seq}}

\newcommand{\Graphs}{\mathcal{G}}

\renewcommand{\Loop}{\textsc{Loop}}
\renewcommand{\Subset}{\textsc{Subset}}
\newcommand{\Indiclique}{\textsc{InClique}}
\newcommand{\Outdiclique}{\textsc{OutClique}}
\newcommand{\Pie}{\textsc{Pie}}
\newcommand{\Dome}{\textsc{Dome++}}
\newcommand{\Reduce}[1]{\stackrel{#1}{\rightarrow}}
\newcommand{\ReduceC}[1]{\stackrel{#1}{\Rightarrow}}

\newcommand{\Abi}{A^{\leftrightarrow}}

\newtheorem{definition}{Definition}

\title{Approaching the Source of Symbol Grounding with Confluent Reductions of Abstract Meaning Representation Directed Graphs}

\author{Nicolas Goulet, Alexandre Blondin Massé \and Moussa Abdenbi \\
  Université du Québec à Montréal \\
  Département d'Informatique \\
  Montréal, Québec, Canada \\
  \texttt{goulet.nicolas@courrier.uqam.ca} }

\begin{document}
\maketitle
\begin{abstract}
Abstract meaning representation (AMR) is a semantic formalism used to represent the meaning of sentences as directed acyclic graphs.
In this paper, we describe how digital dictionaries can be represented as the union of their definitions modeled as AMR digraphs or as plain English. 
These graph representations of dictionaries are then reduced in a \emph{confluent} manner, i.e. in a manner yielding unique results, for the purpose of studying the properties of their structures. 
Finally, the properties of these reduced digraphs are analyzed and discussed in relation to the identification of their respective \textit{minimal grounding sets}, i.e. the smallest set of words which needs to be \textit{grounded} by prior learning in order to define all the rest.

\end{abstract}

\section{Introduction}

When we encounter a word with unknown meaning, we look up its definition.
If, in the definition, we find another unknown word, then we can look it up too, repeating this process until it eventually stops.
However, a language cannot be fully learned from a dictionary via definition look-up alone: \textit{the meaning of some words has to be acquired beforehand}.
Explaining how and why humans have the unique capacity to \textit{ground in arbitrary symbols the things in the world they refer to} in order to break this circularity is known as the symbol grounding problem \cite{harnad1990symbol, harnad2024language}.
One might then ask : how many words -- and which words -- do humans need to learn by means other than dictionary look-up so that all the rest of the words in the dictionary can be defined?
This paper introduces a methodology to probe this question by modeling dictionaries contents as \textit{Abstract Meaning Representation} (AMR) directed graphs (digraphs).

This paper is therefore not about the process of word grounding itself but rather the approach of its source : identifying the set of words that need to be grounded via sensorimotor experience before new words can be indirectly grounded using verbal definitions of previously grounded words.
We propose to call such sets \emph{grounding sets}.
To study and identify grounding sets, there are two sources that can be used : the previously mentioned dictionaries (complete sets of every word in a language with an approximate definition for it) and the \textit{mental lexicons} of human brains. 
The discussion for the latter is a topic for a completely different paper and we therefore focus on the first source. 
This paper also builds upon previously existing methodology \cite{masse2008meaning, vincent2016latent} in the literature that has shown that given a dictionary, identifying a set of words that is sufficient to define all remaining words (i.e., identifying a grounding set) is formally equivalent to identifying a feedback vertex set (FVS) when representing dictionaries as digraphs. 

The paper is organized as follows.
Section \ref{sec:prelim} introduces the necessary definitions and notation.
Section \ref{sec:amr-digraphs} describes our methodology for modeling dictionaries definitions in AMR digraphs.
Section \ref{sec:reductions} presents the digraph reductions we use to \textit{confluently} and \textit{non-confluently} reduce digraphs.
The experiments and their results are described in Section \ref{sec:experi}. We then discuss in Section \ref{sec:discussion} the results before concluding. 


\section{Preliminaries}\label{sec:prelim}

This section introduces the necessary definitions and notation for the following sections.

Subsection~\ref{ss:dictionaries} explains how dictionaries can be formally represented and transformed into directed graphs.
Subsection~\ref{ss:symbol} introduces methodological limitations of previous work in the literature.
Finally, Subsection~\ref{ss:amr} summarizes the main ideas of abstract meaning representation (AMR).

\subsection{Dictionaries}\label{ss:dictionaries}

In its most simple form, a dictionary is simply a set of lexical units with a map associating to each lexical unit a definition.
Some dictionaries have stronger structure than others.
In that spirit, we distinguish two types of dictionaries in this paper, as described by Definitions~\ref{def:raw-dict} and \ref{def:disamb-dict}.
Let $\Strings$ be the set of all strings on a given alphabet.

\begin{definition}\label{def:raw-dict}
    A \emph{raw} dictionary is an ordered pair $D = (L, d)$, where $L$ is a finite set of \emph{lexemes} and $d : L \rightarrow S$ is a map associating each lexeme to a nonempty string called its \emph{raw definition}.
\end{definition}

Although raw dictionaries are useful to human, their structure is quite limited when one wishes to study the definitional relations between lexemes.
A richer representation can be used.
Given a set $A$, let $\Seq(A)$ be the set of all finite sequences of elements taken in $A$.

\begin{definition}\label{def:disamb-dict}
    A \emph{disambiguated} dictionary is an ordered pair $D = (L, d)$, where $L$ is a finite set of \emph{lexemes} and $d : L \rightarrow \Seq(L)$ is a map associating each lexeme to a nonempty sequence of lexemes called its \emph{disambiguated definition}.
\end{definition}

Disambiguated dictionaries can conveniently be represented by a digraph: It suffices to represent each lexeme in $L$ by a vertex, and add an arc between two lexemes $\ell_1$ and $\ell_2$ whenever $\ell_1 \in d(\ell_2)$, i.e. $\ell_1$ is a lexeme occurring in the definition of $\ell_2$.
Unfortunately, to the best of our knowledge, disambiguated dictionaries are not available in practice and must be built using natural language processing tools.
As we show in Section \ref{sec:amr-digraphs}, abstract meaning representation seems to be well adapted to this task.

\subsection{Previous Work}\label{ss:symbol}
Identifying grounding sets and analyzing the structures of English dictionaries has previously been done in the literature \cite{masse2008meaning, vincent2016latent}.
These authors have shown that identifying the smallest grounding set (Minset) of a dictionary modeled as a graph is formally equivalent to identifying the \textit{minimal feedback vertex set} (MFVS), which is the smallest set of nodes that can be removed from a digraph to make it acyclic (see appendix \ref{section:notation} for a formal definition). 
However, these papers only took into consideration first definition of each open-class word in a dictionary. 
This was done to bypass the issue of \textit{word sense disambiguation} : when a word used in a definition itself has more than one definition, how can we \textit{algorithmically} identify which is the intended one?
By keeping only the first definition, the authors proposed to assume that this definition was the intended meaning of the word, whenever it was used to define another word in the dictionary. 
Another approximation was the decision to only consider open-class words and reject close-class words as the authors were more interested in the implication for \textit{category learning} and thus proposed to consider open-class words names of categories. 
A major issue these authors faced was the \textit{computational cost} of identifying Minsets : the graphs built from dictionaries are often too large to be computed in reasonable time, even when using state-of-the-art techniques in MFVS solving.

The papers have also shown that these graphs representations of dictionaries share structural similarities that are of psycholinguistical interests. 
The structures we will consider in this paper are as follows. 
The graph in its initial state is defined as \textit{complete}.
Then, the \textbf{kernel} is obtained by recursively removing both words that are not defined and words that are not used to define any other.
The \textbf{reduced kernel} is what remains of the dictionary after graph reductions have been applied to it.
It is this last structure that is used as input to a MFVS-solver to identify Minsets.


\subsection{Abstract Meaning Representation}\label{ss:amr}

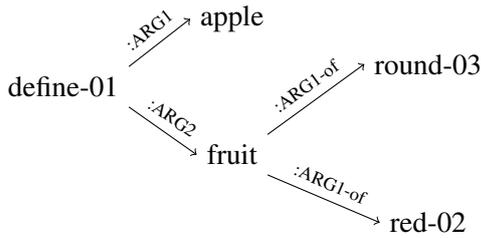
\begin{figure}
    \centering
    \begin{tikzpicture}
    \begin{scope}[scale=0.6]
        \node (d) at (0,0) {$\text{define-01}$};
        \node (a) at (3.7,1.5) {$\text{apple}$};
        \node (f) at (3.7,-1.5) {$\text{fruit}$};
        \node (r) at (8, 0.5)  {$\text{round-03}$};
        \node (r2) at (8,-3) {$\text{red-02}$};
        
        \path[->,draw]
            (d.north east) -- (a.west) node[midway, sloped, above] {\scriptsize \text{:ARG1}};
        \path[->,draw]
            (d.south east) -- (f.west) node[midway, sloped, above] {\scriptsize \text{:ARG2}};
        \path[->,draw]
            (f.north east) -- (r.west) node[midway, sloped, above] {\scriptsize \text{:ARG1-of}};
        \path[->,draw]
            (f.south east) -- (r2.west) node[midway, sloped, above] {\scriptsize \text{:ARG1-of}};
    \end{scope}
\end{tikzpicture}
    \caption{AMR obtained from translating the sentences "apple is defined as a red round fruit" and "a red round fruit is the definition of apple". As these sentences are definitions, \textit{define-01} is the focus of the AMR. In Propbank, the ARG1 of define-01 is the \textit{thing defined} and the ARG2 is the \textit{definition}.}
    \label{fig:apple_def}
\end{figure}

Abstract meaning representation (AMR) is a semantic formalism designed to represent the \textit{meaning} of a given sentence, whilst abstracting away from syntactic idiosyncrasies \cite{banarescu2013abstract}.
In this formalism, sentences are represented by rooted, directed acyclic graphs where vertices represent \emph{concepts} and labelled arcs are used to represent \emph{semantic relationships} between them.
Those concepts are often lifted from Propbank \cite{palmer2005proposition}, a dataset with disambiguated concepts and arguments, which are specific relationships a given concept can have with other concepts.
An interesting property of AMR is that different sentences with the same meaning should have the same AMR.
For example, the sentences ``apple is defined as a red round fruit'' and ``a red round fruit is the definition of apple'' share the AMR representation shown in Figure \ref{fig:apple_def}. AMR has been of help for a variety of use-cases such as text summarization \cite{dohare2017text, kouris2024text}, machine translation \cite{li2022improving} and grounded human-robot conversations\cite{bonial2023abstract}.
For a more complete review of AMR applications, refer to \cite{tohidi2022short}.

This semantic formalism is of interest for embedding definitions for two main reasons. First is the potential for word-sense disambiguation. In Propbank, words with multiple meanings are assigned numbered senses to distinguish between each meaning. For example, the word \emph{admonish} has two distinct frames, each associated with specific arguments.
On one hand, admonish-01, whose meaning is \emph{persuade warningly}, takes 3 arguments: the persuader (ARG0), the persuaded agent (ARG1) and the persuaded action (ARG2).
On the other hand, admonish-02, whose meaning is \emph{chastise} takes 3 arguments as well, but with different meaning: the chastiser (ARG0), the chastised (ARG1) and the wrongdoing (ARG2).

In the previous example, disambiguation could be accomplished in two manners. The first one is via the numbered labels : when facing multiple definitions for a given word, assign when possible a unique numbered label to represent specific meaning. The other option is by translating it to another defined concept entirely. AMR is like translating from English to a different language entirely: Some words will be represented by different arbitrary symbols. 

The second point of interest for embedding definitions in AMR is enrichment of the structure via a less atomic representation of meaning. For example while a word like \textit{negation} is defined in a dictionary, AMR represents the concept of \textit{negation} using polarity, or the notion of \textit{conjunction} through instances like \textit{or}. This can also be described as moving from a less strictly \textit{lexicographic} approach to identifying \textit{grounding sets} to a more \textit{semantic} one.


\section{From Dictionaries to AMR Digraphs}\label{sec:amr-digraphs}

We now describe the process of embedding definitions of digital dictionaries into AMR.
Subsection \ref{ss:format} introduces the desired structure when translating a definition into AMR.
Subsection \ref{ss:disamb} in turn describes how AMR is leveraged to address polysemy.
Subsection \ref{ss:amr-digraphs} completes the construction of AMR digraphs.

\subsection{AMR Definitional Digraphs}\label{ss:format}

\begin{table}
\centering
\begin{tabular}{ll}
\hline
1. ``$s$ is defined as $d$.'' \\
2. ``The definition of $s$ is $d$.'' \\
3. ``$s$ has for definition $d$''. \\
4. ``$d$ is the definition of $s$.'' \\
5. ``$s$ is defined by $d$.'' \\
6. ``$s$ gets defined as $d$.'' \\
\hline
\end{tabular}
\caption{Some possible rephrasing of definitions, highlighting the definitional relation.}
\label{tab:sentences}
\end{table}

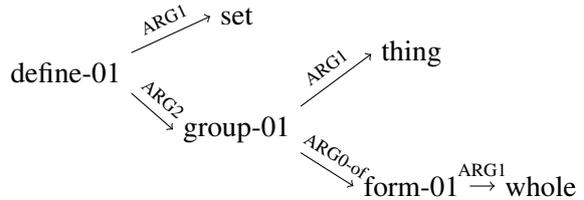
\begin{figure}
    \centering
    \begin{tikzpicture}
    \begin{scope}[scale=0.5]
        \node (d) at (-0.5,0) {$\text{define-01}$};
        \node (s) at (4,1.5) {$\text{set}$};
        \node (g) at (4,-1.5) {$\text{group-01}$};
        \node (t) at (8.6, 0.5)  {$\text{thing}$};
        \node (f) at (8.6,-3) {$\text{form-01}$};
        \node (w) at (12, -3) {$\text{whole}$};
        
        \path[->,draw]
            (d.north east) -- (s.west) node[midway, sloped, above] {\scriptsize \text{ARG1}};
        \path[->,draw]
            (d.south east) -- (g.west) node[midway, sloped, above] {\scriptsize \text{ARG2}};
        \path[->,draw]
            (g.north east) -- (t.west) node[midway, sloped, above] {\scriptsize \text{ARG1}};
        \path[->,draw]
            (g.south east) -- (f.west) node[midway, sloped, above] {\scriptsize \text{ARG0-of}};
        \path[->,draw]
            (f.east) -- (w.west) node[midway, sloped, above] {\scriptsize \text{ARG1}};
    \end{scope}
\end{tikzpicture}
    \caption{AMR generated from the sentence ``set is defined as a group of things that form a whole''.}
    \label{fig:set_def}
\end{figure}

The first step to embed the content of a dictionary into the AMR formalism is to create an AMR digraph out of each definition from the dictionary.
In particular, such an approach needs to capture the \emph{definitional} relation between the \emph{defining words} and the \emph{defined word}.
In that spirit, one might rephrase each definition according to any of the sentences given in Table~\ref{tab:sentences}, where $s$ is the defined symbol and $d$ is the given definition of symbol $s$.
Such sentences can then be translated into AMR embeddings using a state-of-the-art sentence-to-graph (StoG) parser \cite{jascob_amrlib_models_2023}.
In most cases, the resulting AMR digraph has the structure shown in Figures \ref{fig:apple_def} and \ref{fig:set_def}, i.e. the root of the graph is ``define-01'', its first argument is $s$ and its second argument points to an AMR digraph of $d$.
An AMR digraph having this structure is called \emph{valid}.

\begin{figure}[tb]
    \centering
    \begin{minipage}{0.225\textwidth}
        \centering
        \resizebox{\textwidth}{!}{\begin{tikzpicture}
    \begin{scope}[scale=0.5]
        \node (d) at (-0.5,0) {$\text{define-01}$};
        \node (w) at (4,1.5) {$\text{way}$};
        \node (s2) at (4,-0.5) {$\text{somehow}$};
        \node (k) at (8, 3)  {$\text{know}$};  
        \node (s) at (8, 1) {$\text{state-01}$};
        \node[draw, rectangle] (p1) at (12, 3) {$-$}; 
        \node[draw, rectangle] (p2) at (12, 1) {$-$};
        
        \path[->,draw]
            (d.north east) -- (w.west) node[midway, sloped, above] {\scriptsize \text{ARG2}};
        \path[->,draw]
            (d.south east) -- (s2.west) node[midway, sloped, above] {\scriptsize \text{manner}};
        \path[->,draw]
            (w.north east) -- (k.west) node[midway, sloped, above] {\scriptsize \text{ARG1-of}};
        \path[->,draw]
            (w.south east) -- (s.west) node[midway, sloped, above] {\scriptsize \text{ARG-of}};
        \path[->,draw]
            (k.east) -- (p1.west) node[midway, sloped, above] {\scriptsize \text{polarity}};
        \path[->,draw]
            (s.east) -- (p2.west) node[midway, sloped, above] {\scriptsize \text{polarity}};
    \end{scope}
\end{tikzpicture}}
        \caption*{(A)}
    \end{minipage}
    \hfill
    \begin{minipage}{0.225\textwidth}
        \centering
        \resizebox{\textwidth}{!}{\begin{tikzpicture}
    \begin{scope}[scale=0.5]
        \node (d) at (-0.5,0) {$\text{define-01}$};
        \node (h) at (4,1.5) {$\text{hereto}$};
        \node (o) at (4,-1.5) {$\text{or}$};
        \node (m) at (8, 0.5)  {$\text{matter}$};   
        \node (d2) at (8, -2.5) {$\text{document}$}; 
        \node (t) at (12, -1.5) {$\text{this}$};
        
        \path[->,draw]
            (d.north east) -- (h.west) node[midway, sloped, above] {\scriptsize \text{ARG1}};
        \path[->,draw]
            (d.south east) -- (o.west) node[midway, sloped, above] {\scriptsize \text{topic}};
        \path[->,draw]
            (o.north east) -- (m.west) node[midway, sloped, above] {\scriptsize \text{op1}};
        \path[->,draw]
            (o.south east) -- (d2.west) node[midway, sloped, above] {\scriptsize \text{op2}};
        \path[->,draw]
            (d2.east) -- (t) node[midway, sloped, above] {\scriptsize \text{mod}};
        \path[->,draw]
            (m.east) -- (t) node[midway, sloped, above] {\scriptsize \text{mod}};
    \end{scope}
\end{tikzpicture}}
        \caption*{(B)}
    \end{minipage}
    \vspace{1em}
    \begin{minipage}{0.225\textwidth}
        \centering
        \resizebox{\textwidth}{!}{\begin{tikzpicture}
    \begin{scope}[scale=0.5]
        \node (c) at (-0.5,0) {\scriptsize $\text{contrast}$};
        \node (d) at (4,1.5) {\scriptsize $\text{define-01}$};
        \node (d2) at (4,-1.5) {\scriptsize $\text{define-01}$};
        \node (o) at (8, 3) {\scriptsize $\text{occasion-02}$};  
        \node (s) at (8, 0.5) {\scriptsize $\text{sometimes}$};
        \node (t) at (8, -1.5) {\scriptsize $\text{time}$}; 
        \node[draw, rectangle] (p) at (8, -3) {\scriptsize $-$};
        \node (p2) at (12, 3) {\scriptsize $\text{particular}$};
        \node (a) at (12, -1.5) {\scriptsize $\text{all}$};
        \path[->,draw]
            (c.north east) -- (d.west) node[midway, sloped, above] {\tiny \text{ARG1}};
        \path[->,draw]
            (c.north east) -- (d2.west) node[midway, sloped, above] {\tiny \text{ARG2}};
        \path[->,draw]
            (d.south east) -- (s.north west) node[midway, sloped, above] {\tiny \text{ARG1}};
        \path[->,draw]
            (d.north east) -- (o.west) node[midway, sloped, above] {\tiny \text{time}};
        \path[->,draw]
            (d2.north east) -- (s.south west) node[midway, sloped, above] {\tiny \text{ARG1}}; 
        \path[->,draw]
            (d2.east) -- (t.west) node[midway, sloped, above] {\tiny \text{time}};
        \path[->,draw, shorten >= 3pt] 
            (d2.south east) -- (p.west) node[midway, sloped, above] {\tiny \text{polarity}};
        \path[->,draw]
            (o.east) -- (p2.west) node[midway, sloped, above] {\tiny \text{mod}};
        \path[->,draw]
            (t.east) -- (a.west) node[midway, sloped, above] {\tiny \text{mod}};
    \end{scope}
\end{tikzpicture}}
        \caption*{(C)}
    \end{minipage}
    \hfill
    \begin{minipage}{0.225\textwidth}
        \centering
        \resizebox{\textwidth}{!}{\begin{tikzpicture}
    \begin{scope}[scale=0.5]
        \node (d) at (-0.5,0) {\scriptsize $\text{define-01}$};
        \node (g) at (3,1.5) {\scriptsize $\text{gov-org}$};
        \node (b) at (3,-1.5) {\scriptsize $\text{building}$};
        \node (n) at (6, 1.5) {\scriptsize $\text{name}$};  
        \node (g2) at (10, 1.5) {\scriptsize $\text{"Guardhouse"}$};
        \node (s) at (6, -1.5) {\scriptsize $\text{soldier}$}; 
        \node (p) at (9, -1.5) {\scriptsize $\text{protect-01}$};
        \node (p2) at (12.5, -1.5) {\scriptsize $\text{place}$};
        \path[->,draw]
            (d.north east) -- (g.west) node[midway, sloped, above] {\tiny \text{ARG1}};
        \path[->,draw]
            (d.south east) -- (b.west) node[midway, sloped, above] {\tiny \text{ARG2}};
        \path[->,draw]
            (g.east) -- (n.west) node[midway, sloped, above] {\tiny \text{name}};
        \path[->,draw]
            (n.east) -- (g2.west) node[midway, sloped, above] {\tiny \text{op1}};
        \path[->,draw]
            (b.east) -- (s.west) node[midway, sloped, above] {\tiny \text{beneficiary}};
        \path[->,draw]
            (s.east) -- (p.west) node[midway, sloped, above] {\tiny \text{ARG0-of}};
        \path[->,draw]
            (p.east) -- (p2.west) node[midway, sloped, above] {\tiny \text{ARG1}};
    \end{scope}
\end{tikzpicture}}
        \caption*{(D)}
    \end{minipage}
    \caption{(A) An AMR without an ARG1. The arc here is mistakenly labeled as \textit{manner}. (B) An AMR without an ARG2. The arc here is mistakenly labeled as \textit{topic}. (C) An AMR with an undesired root node. The model interpreted the definition as a contrast between two definitions. (D) An AMR with a symbol defined by a subgraph.}
    \label{fig:error_cases}
\end{figure}

Unfortunately, in many cases, the resulting graph does not exhibit the desired structure.
Figure \ref{fig:error_cases} provides examples of such \emph{invalid} cases.
If none of the rephrasing proposed in Table~\ref{tab:sentences} is able to yield a valid AMR digraph, in some cases, it is possible to partially mitigate the loss by patching the resulting digraph.
More specifically, assume that the produced AMR digraph for symbol $s$ and definition $d$ has a valid root node and has a single node designated by ARG1, which is $s$.
Then it is possible to delete everything but the root and $s$, generate a new AMR for $d$, and then bind to ARG2 the root of the newly generated AMR digraph, to obtain a valid AMR digraph.
This process is illustrated in Figure \ref{fig:error_corr}.

\begin{figure}
    \centering
    \begin{minipage}{0.225\textwidth}
        \centering
        \resizebox{\textwidth}{!}{\begin{tikzpicture}
    \begin{scope}[scale=0.5]
        \node (d) at (-0.5,0) {$\text{define-01}$};
        \node (w) at (4,1.5) {$\text{wacky}$};
        \node (s) at (4,0) {$\text{silly}$};
        \node (o) at (4, -1.5)  {$\text{or}$};
        \node (e) at (8, -0.5) {$\text{excite}$};
        \node (a) at (8, -2) {$\text{amuse-01}$};
        
        \path[->,draw]
            (d.north east) -- (w.west) node[midway, sloped, above] {\scriptsize \text{ARG1}};
        \path[->,draw]
            (d.east) -- (s.west) node[midway, sloped, above] {\scriptsize \text{ARG2}};
        \path[->,draw]
            (d.south east) -- (o.west) node[midway, sloped, above] {\scriptsize \text{manner}};
        \path[->,draw]
            (o.north east) -- (e.west) node[midway, sloped, above] {\scriptsize \text{op1}};
        \path[->,draw]
            (o.east) -- (a.west) node[midway, sloped, above] {\scriptsize \text{op2}};
    \end{scope}
\end{tikzpicture}}
        \caption*{(1)}
    \end{minipage}
    \hfill
    \begin{minipage}{0.225\textwidth}
        \centering
        \resizebox{\textwidth}{!}{\begin{tikzpicture}
    \begin{scope}[scale=0.5]
        \node (d) at (-0.5,0) {$\text{define-01}$};
        \node (w) at (2,2) {$\text{wacky}$};
        \node (s) at (4,0) {$\text{silly}$};
        \node (o) at (6, 0)  {$\text{or}$};
        \node (e) at (8, 2) {$\text{excite}$};
        \node (a) at (8, -2) {$\text{amuse-01}$};
        
        \path[->,draw]
            (d.north east) -- (w.south) node[midway, sloped, above] {\scriptsize \text{ARG1}};
        \path[->,draw]
            (o.north east) -- (e.south) node[midway, sloped, above] {\scriptsize \text{op1}};
        \path[->,draw]
            (o.east) -- (a.north) node[midway, sloped, above] {\scriptsize \text{op2}};
        \path[->, draw]
            (s.east) -- (o.west) node[midway, sloped, above]{\scriptsize \text{manner}};
    \end{scope}
\end{tikzpicture}}
        \caption*{(2)}
    \end{minipage}
    \vspace{1em}
    \begin{minipage}{0.45\textwidth}
        \centering
        \resizebox{\textwidth}{!}{\begin{tikzpicture}
    \begin{scope}[scale=0.5]
        \node (d) at (-0.5,0) {$\text{define-01}$};
        \node (w) at (4,1.5) {$\text{wacky}$};
        \node (s) at (4,-1.5) {$\text{silly}$};
        \node (o) at (8, -1.5)  {$\text{or}$};
        \node (e) at (12, -0.5) {$\text{excite}$};
        \node (a) at (12, -2) {$\text{amuse-01}$};
        
        \path[->,draw]
            (d.north east) -- (w.west) node[midway, sloped, above] {\scriptsize \text{ARG1}};
        \path[->,draw]
            (d.south east) -- (s.west) node[midway, sloped, above] {\scriptsize \text{ARG2}};
        \path[->,draw]
            (s.east) -- (o.west) node[midway, sloped, above] {\scriptsize \text{manner}};
        \path[->,draw]
            (o.north east) -- (e.west) node[midway, sloped, above] {\scriptsize \text{op1}};
        \path[->,draw]
            (o.east) -- (a.west) node[midway, sloped, above] {\scriptsize \text{op2}};
    \end{scope}
\end{tikzpicture}}
        \caption*{(3)}
    \end{minipage}
    \caption{Process of correcting an invalid AMR digraph obtained by SToG for the sentence ``wacky is defined as silly in an exciting or amusing way''. \textbf{(1)} An invalid AMR digraph is detected. \textbf{(2)} Everything but the valid parts are deleted ; a new AMR digraph is generated using only the definition (``silly in an exciting or amusing way''). \textbf{(3)} The AMR digraph of the definition is bound to the root label by an ARG2 labeled arc.}
    \label{fig:error_corr}
\end{figure}

\subsection{Handling Polysemy}\label{ss:disamb}
Once the number of valid AMRs has been maximized, there remains the issue of \textit{polysemy}: There can be many AMR definitional digraphs having the same label for \textit{s}, since there is no guarantee that the polysemy of the processed digital dictionary is aligned with the polysemy represented in the AMR space.

In that case, we can rely on a strategy similar to that used in \cite{vincent2016latent}, which consists in keeping only the first meaning with respect to the order used in the source digital dictionary.
This process must be extended across all symbols to take into account collisions, as AMR might translate different words using the same label, and to prevent alteration of the structure of the dictionary.

\subsection{AMR Digraphs}\label{ss:amr-digraphs}

To finalize the transformation of the AMR digraphs, it remains to remove the AMR arcs and add arcs in a way that the definitional relation is correctly reflected. 
Since at that point each AMR is rooted with the label \textit{define-01}, has a single AMR node to represent $s$ and has a single subgraph to represent $d$, it can be assumed that each node in the definitional subgraph is used to define the word at the end of ARG1.
The root node can then be safely deleted and an outgoing arc labeled \textit{define-01} is added to each node of the definitional subgraph towards the defined word.

This process is necessary to align the AMR formalism with the previously described approach to the \textit{symbol grounding problem}.
Figure \ref{fig:wacky_comparison} illustrates the difference between the two definition graph modeling approaches.
We claim that the arcs added via this bypass capture \textit{definitional relationships} similar to those used in \cite{vincent2016latent}.
As consequence, we can benefit from the same graph theoretical representation to identify \textit{grounding sets} from AMR digraphs. We also preserve the removed AMR arcs to be added back to the \textit{grounding sets}.
More formally, \textit{AMR digraphs} of dictionaries are defined as the union of the individual AMR definitional digraphs created from the embedding of the definitions described in the last subsections.
This union can be seen as the graphical representation of the contents of a digital dictionary translated into AMR embeddings.

 \begin{figure}
    \centering
    \begin{minipage}{0.22\textwidth}
        \centering
        \begin{tikzpicture}
    \begin{scope}[scale=0.5]
        \node (w) at (0,0) {\footnotesize $\text{wacky}$};
        \node (s) at (-3,-1.5) {\footnotesize $\text{silly}$};
        \node (o) at (-1, -1.5)  {\footnotesize $\text{or}$};
        \node (e) at (3, -0.8) {\footnotesize $\text{excite}$};
        \node (a) at (3, -2.2) {\footnotesize $\text{amuse-01}$};
        

        \path[->,draw]
            (s.east) -- (o.west) node[midway, sloped, above] {\scriptsize \text{manner}};
        \path[->,draw]
            (o.north east) -- (e.west) node[midway, sloped, above] {\scriptsize \text{op1}};
        \path[->,draw]
            (o.east) -- (a.west) node[midway, sloped, above] {\scriptsize \text{op2}};
        
        \path[->,draw,blue]
            (s) edge[out=45, in=160] (w);
        \path[->,draw,blue]
            (o) edge[out=90, in=250] (w);
        \path[->,draw,blue]
            (e) edge[out=120, in=360] (w);
        \path[->,draw,blue]
            (a) edge[out=150, in=350] (w);
    \end{scope}
\end{tikzpicture}
        \caption*{(A)}
    \end{minipage}
    \hfill
    \begin{minipage}{0.22\textwidth}
        \centering
        \begin{tikzpicture}
    \begin{scope}[scale=0.5]
        \node (w) at (0,0) {\footnotesize $\text{wacky}$};
        \node (s) at (-3,-2.2) {\footnotesize $\text{silly}$};
        \node (e) at (-1, -2.2)  {\footnotesize $\text{excite}$};
        \node (a) at (1, -2.2) {\footnotesize $\text{amuse}$};
        \node (w2) at (3, -2.2) {\footnotesize $\text{way}$};
        
        \path[->,draw]
            (s) -- (w) node[midway, sloped, above] {} [out=30,in=210];
        \path[->, draw]
            (e) -- (w) node[midway, sloped, above] {} [out=60,in=240];
        \path[->, draw]
            (a) -- (w) node[midway, sloped, above] {} [out=120,in=300];
        \path[->, draw]
            (w2) -- (w) node[midway, sloped, above] {} [out=150,in=330];
    \end{scope}
\end{tikzpicture}
        \caption*{(B)}
    \end{minipage}
    \caption{\textbf{(A)} An AMR where the root label has been bypassedet . Each blue arc has been added after this bypass and is labelled \textit{define-01} to signify a definitional relationship. \textbf{(B)} is the previous definitional graph format used in \cite{vincent2016latent}. Both are made from the definition of "wacky".}
    \label{fig:wacky_comparison}
\end{figure}


\section{Reducing AMR Digraphs}\label{sec:reductions}

A common first step of many algorithms solving the minimum feedback vertex set problem consists in applying reductions to the original digraph, while preserving, in some sense, its MFVS.
In this section, we recall some of those reductions and discuss their \emph{confluence property}.

\subsection{Digraph Reductions}

Let $\Graphs$ be the set of all finite digraphs.
A map $T : \Graphs \rightarrow \Graphs$ is called a \emph{reduction} if, for any $G, G' \in \Graphs$, such that $G = (V, A)$, $G' = (V', A')$ and $G' = T(G)$, either $|V'| < |V|$, or $|V'| = |V|$ and $|A'| < |A|$.
In other words, a reduction either remove at least one vertex or, in the case where no vertex is removed, it removes at least one arc.
Let $T$ be a reduction, $G, G' \in \Graphs$ be such that $G' = T(G)$ and $U \subseteq V(G)$.
We say that $T$ is \emph{MFVS-preserving for $G$, with respect to $U$} if $U \cap V' = \emptyset$ and, for any MFVS $U'$ of $G'$, the set $U' \cup U$ is a MFVS of $G$.
The set $U$ is called a \emph{partially constructed solution from $T$}.

Many MFVS-preserving reductions have been proposed in the literature \cite{levy_1988,lin_2000,lemaic_2008,kiesel_2022}.
We recall some of them below.
Let $G = (V, A)$ be any directed graph, $u, v \in V$.
We define the following predicates:

\begin{itemize}
  \item $\ell(G, u)$ if $(u,u) \in A$;
  \item $c(G, U)$ if $U$ is a diclique of $G$; 
  \item $i(G, u)$ if $\neg \ell(G, u)$ and $c(G, N^-_G(u))$;
  \item $o(G, u)$ if $\neg \ell(G, u)$ and $c(G, N^+_G(u))$;
  \item $s(G, u, v)$ if $\neg \ell(G,v)$, $(u,v) \in A$, $(v,u) \in A$, $N^-_G(v) \subseteq N^-_G(u)$ and $N^+_G(v) \subseteq N^+_G(u)$;
  \item $p(G, u, v)$ if $(u,v)$ is acyclic in $G - \Abi(G)$.
  \item $d(G, u, v)$ if $(u,v) \in A$ and any $vu-path$ in $G - \Abi(G)$ goes through a vertex in $N_{G - \Abi(G)}^-(v) \cup N_{G - \Abi(G)}^+(u)$.
\end{itemize}

Pointed reductions, i.e. reductions targeting a specific vertex or arc, can then be defined as follows:

\footnotesize
\begin{eqnarray*}
  \Loop(G, u) & = & \begin{cases}
    G - u & \mbox{if $\ell(G, u)$;} \\
    G & \mbox{otherwise.}
  \end{cases} \\
  \Indiclique(G, u) & = & \begin{cases}
    G \circ u & \mbox{if $i(G, u)$;} \\
    G & \mbox{otherwise.}
  \end{cases} \\
  \Outdiclique(G, u) & = & \begin{cases}
    G \circ u & \mbox{if $o(G, u)$;} \\
    G & \mbox{otherwise.}
  \end{cases} \\
  \Subset(G, u, v) & = & \begin{cases}
    G - u & \mbox{if $s(G, u, v)$;} \\
    G & \mbox{otherwise.}
  \end{cases} \\
  \Pie(G, u, v) & = & \begin{cases}
    G - (u,v) & \mbox{if $p(G, u, v)$;} \\
    G & \mbox{otherwise.}
  \end{cases} \\
  \Dome(G, u, v) & = & \begin{cases}
    G - (u,v) & \mbox{if $d(G, u, v)$;} \\
    G & \mbox{otherwise.}
  \end{cases}
\end{eqnarray*}
\normalsize

\begin{figure}
    \centering
    \begin{tikzpicture}
    \begin{scope}[scale=0.6]
        \node (u) at (0,0) {$u$};
            \node (p1) at ($ (u) + (160:3.3) $) {$p_1$};
            \node (p2) at ($ (u) + (180:1.75) $) {$p_2$};
            \node (p3) at ($ (u) + (-160:3.3) $) {$p_3$};
            \node (s1) at ($ (u) + (35:1.9) $) {$s_1$};
            \node (s2) at ($ (u) + (-35:1.9) $) {$s_2$};
                
        \path[->,draw]
            (u) edge [] node {} (s1)
            (u) edge [] node {} (s2)
            (p2) edge [] node {} (u);
        \path[->,draw,bend left = 17]
            (p1) edge [] node {} (p2)
            (p2) edge [] node {} (p1)
            (p3) edge [] node {} (p2)
            (p2) edge [] node {} (p3)
            (p1) edge [] node {} (p3)
            (p3) edge [] node {} (p1);
        \path[->,draw,bend left]
            (p1) edge [] node {} (u);
        \path[->,draw,bend right]
            (p3) edge [] node {} (u);

        \node (up) at (7.5,0) {};
            \node (p1p) at ($ (up) + (160:3.3) $) {$p_1$};
            \node (p2p) at ($ (up) + (180:1.75) $) {$p_2$};
            \node (p3p) at ($ (up) + (-160:3.3) $) {$p_3$};
            \node (s1p) at ($ (up) + (35:1.9) $) {$s_1$};
            \node (s2p) at ($ (up) + (-35:1.9) $) {$s_2$};
                
        \path[->,draw]
            (p2p) edge [] node {} (s1p)
            (p2p) edge [] node {} (s2p);
        \path[->,draw,bend left = 17]
            (p1p) edge [] node {} (p2p)
            (p2p) edge [] node {} (p1p)
            (p3p) edge [] node {} (p2p)
            (p2p) edge [] node {} (p3p)
            (p1p) edge [] node {} (p3p)
            (p3p) edge [] node {} (p1p);
        \path[->,draw,bend right]
            (p3p) edge [] node {} (s2p)
            (p3p) edge [] node {} (s1p);
        \path[->,draw,bend left]
            (p1p) edge [] node {} (s1p)
            (p1p) edge [] node {} (s2p);
    \end{scope}
\end{tikzpicture}
    \caption{Illustration of the \Indiclique~reduction. 
    Consider on the left the subgraph above of a digraph $G$ where $i(G, u)$ is true, indeed $N^-_G(u) = \{p_1, p_2, p_3 \}$ form a diclique.
    Then $\Indiclique(G,u)$ is applicable and on the right we see the result.}
    \label{fig:indiclique}
\end{figure}
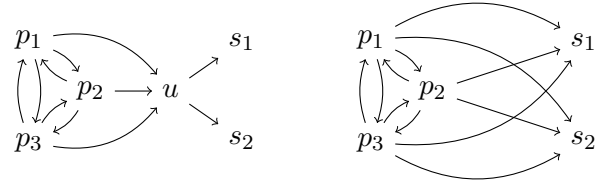

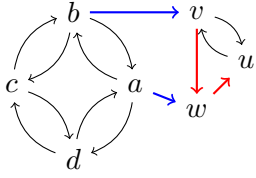
\begin{figure}
    \centering
    \begin{tikzpicture}[highred/.style={thick, draw=red}, highblue/.style={thick, draw=blue, fill=blue!30}]
    \begin{scope}[scale=0.65]
        \node (x) at (-1.5,0) {};
            \node (a) at ($ (x) + (0:1.25) $) {$a$};
            \node (b) at ($ (x) + (90:1.5) $) {$b$};
            \node (c) at ($ (x) + (180:1.25) $) {$c$};
            \node (d) at ($ (x) + (-90:1.5) $) {$d$};
        \node (y) at (1,0.5) {};
            \node (u) at ($ (y) + (0:1) $) {$u$};
            \node (v) at ($ (y) + (90:1) $) {$v$};
            \node (w) at ($ (y) + (-90:1) $) {$w$};
                
        \path[->,draw,highblue]
            (a) edge [] node {} (w)
            (b) edge [] node {} (v);
        \path[->,draw,highred]
            (w) edge [] node {} (u)
            (v) edge [] node {} (w);
        \path[->,draw,bend left]
            (a) edge [] node {} (b)
            (b) edge [] node {} (c)
            (c) edge [] node {} (d)
            (d) edge [] node {} (a)
            (v) edge [] node {} (u)
            (u) edge [] node {} (v)
            (b) edge [] node {} (a)
            (c) edge [] node {} (b)
            (d) edge [] node {} (c)
            (a) edge [] node {} (d);
    \end{scope}
\end{tikzpicture}
    \caption{Illustration of the \Pie~reduction. 
    Consider the digraph $G$ above.
    Then \Pie~is applicable on the blue and red arcs. 
    Indeed, there no circuit going through the blue arcs $(b,v)$ and $(a,w)$ in $G$.
    The same is true for the red arcs $(v,w)$ and $(w,u)$ in $G - \Abi(G)$. 
    Therefore we can remove the blue and red arcs from $G$.}
    \label{fig:pie}
\end{figure}

The effect of the reductions $\Indiclique$ and $\Pie$ are illustrated respectively in Figures \ref{fig:indiclique} and \ref{fig:pie}.
Moreover, proofs that each of these reductions are MFVS-preserving can be found in \cite{levy_1988} for $\Loop$, in Lemaic's thesis for $\Indiclique$ and $\Outdiclique$ \cite{lemaic_2008}, in Lin and Jou's paper for $\Pie$ \cite{lin_2000}, and in Kiesel and Schidler's paper for $\Subset$ and $\Dome$ \cite{kiesel_2022}.

\subsection{Confluence}

Let $G$ be a digraph and $\mathcal{R}$ be a set of reductions.
We say that $G$ is \emph{$\mathcal{R}$-irreducible} if, for any $R \in \mathcal{R}$, $R(G) = G$.
If there exists a sequence of reductions $R_1, R_2, \ldots, R_k \in \mathcal{R}$ such that $G' = (R_k \circ \cdots R_2 \circ R_1)(G)$, then we write $G \Reduce{\mathcal{R}} G'$ and say that \emph{$G$ reduces to $G'$}.
If $G'$ is $\mathcal{R}$-irreducible, then we write $G \ReduceC{\mathcal{R}} G'$ and then we say that \emph{$G$ fully reduces to $G'$}.
Finally, we say that $\mathcal{R}$ is \emph{confluent} if, for any digraph $G$, the conditions $G \ReduceC{\mathcal{R}} G'$ and $G \ReduceC{\mathcal{R}} G''$ imply $G' = G''$.
Roughly speaking, confluence means that a digraph always fully reduces to the same graph, whatever the order in which we apply the reductions.

It was recently proved that the pointed operators $\Loop(\cdot, \cdot)$, $\Indiclique(\cdot, \cdot)$, $\Outdiclique(\cdot, \cdot)$, $\Subset(\cdot, \cdot, \cdot)$ and $\Pie(\cdot, \cdot, \cdot)$ form a set of confluent reductions \cite{Abdenbi_2024}.
In particular, this means that the non pointed versions $\Loop(G)$, $\Indiclique(G)$, $\Outdiclique(G)$, $\Subset(G)$ and $\Pie(G)$ of the operators acting on all vertices or arcs yield the same result, whatever the order in which the vertices or the arcs are traveled.
More generally, once the confluence of a set of reductions $\mathcal{R}$ is established, we are free to apply the confluent reductions in any order, with the guarantee that the final irreducible graph is unique up to isomorphism. 
From an algorithmic and computational complexity perspective, it is more efficient to apply the least costly reductions first, typically those with local applicability criteria, i.e., those involving a single vertex or arc and its neighborhood, such as $\Loop$ or $\Subset$. 
Afterward, we apply those that require more computation time, typically those with graph traversal-based criteria, such as $\Pie$ or $\Dome$. 

\begin{algorithm}
    \caption{Reduce with a set $\mathcal{R}$ of reductions}\label{algo:reduce-conf}
    \hspace*{\algorithmicindent} \textbf{\textit{Input:}} $G$ : digraph, $\mathcal{R}$: set of reductions,\\
    \hspace*{\algorithmicindent} $\rho$ : priority function\\
    \hspace*{\algorithmicindent} \textbf{\textit{Output:}} an \emph{$\mathcal{R}$-irreducible} digraph 
    \begin{algorithmic}[1]
        \Function{Reduced}{$G$, $\mathcal{R}$, $\rho$}
            \State $m \gets \max\{\rho(R) \mid R \in \mathcal{R}\}$
            \State \Return \Call{Reduced}{$G$, $\mathcal{R}$, $\rho$, $m$}
        \EndFunction
        \Function{Reduced}{$G$, $\mathcal{R}$, $\rho$, $p$}
            \If{$p > 0$}
                \State $G' \gets G$
                \Repeat
                    \State $G'' \gets G'$
                    \State $G'' \gets \Call{Reduced}{G'', \mathcal{R}, \rho, p - 1}$
                    \For{$R \in \mathcal{R}$}
                        \If{$\rho(R) = p$}
                            \State $G'' \gets R(G'')$
                        \EndIf
                    \EndFor
                \Until{$G' = G''$}
                \State \Return $G'$
            \Else
                \State \Return $G$
            \EndIf
        \EndFunction
    \end{algorithmic}
\end{algorithm}\label{algo:reductions}

Algorithm \ref{algo:reductions} can be used to reduce a digraph with a given set of allowed reductions $\mathcal{R}$ whose order of application is driven by a priority function $\rho : \mathcal{R} \rightarrow \mathbb{N}_{> 0}$ that associates with each reduction a positive integer (the smaller the value, the higher the priority).
For instance, by setting $\mathcal{R}_c = \{\Loop, \Subset, \Indiclique, \Outdiclique, \Pie\}$ and
\[\rho_c(R) = \begin{cases}
  1, & \mbox{if $R \in \{ \Loop, \Subset,$} \\
  & \mbox{\hphantom{if~}$\Indiclique, \Outdiclique \}$;} \\
  2, & \mbox{if $R = \Pie$,}
\end{cases}\]
we obtain an efficient and confluent reduction algorithm $A_{\mathrm{c}}$.
Similarly, by setting $\mathcal{R}_{nc} = \mathcal{R}_c \cup \{\Dome\}$, $\rho_{nc}(\Dome) = 3$ and $\rho_{nc}(R) = \rho_c(R)$ for $R \neq \Dome$, we obtain another reduction algorithm $A_{nc}$, which has the potential to reduce the digraph further more, to the cost of sacrificing confluence.

\section{Experimental Results}\label{sec:experi}

The section details the carried out experiment on real digital dictionaries.

We assembled $8$ datasets coming from 5 different sources. 
Two of the dictionaries, the Longman’s Dictionary of Contemporary English (LDOCE) \cite{procter_1978}, and the Cambridge International Dictionary of English (CIDE) \cite{procter_1995}, are built using a so-called \emph{controlled vocabulary}, i.e. the words used in the definitions are limited as much as possible. 
LDOCE is an advanced learner’s dictionary, originally published in 1978, while CIDE is a dictionary originally developed in 1995 for advanced learners of English using the Cambridge Corpus. 
The 3rd dictionary is the 11th edition of the Merriam-Webster’s Collegiate Dictionary (MWC), published in 2003 \cite{merriam_2003}. 
The next dictionaries come from Wordsmyth \cite{wordsmyth_2017}, a linguistical educational project.
They are divided into four specialized dictionaries: the Wordsmyth Educational Dictionary-Thesaurus (WEDT), first developed in 1980, followed later by the Wordsmyth Learner’s Dictionary-Thesaurus (WLDT), the Wordsmyth Children’s Dictionary-Thesaurus (WCDT) and the Wordsmyth Illustrated Learner’s Dictionary (WILD). 
The first two are targeted at adults, WEDT being for advanced learners and WLDT for beginners, while the last two are aimed at children. 
Finally, WordNet (WN) \cite{fellbaum_1998} is a well-known lexical network, whose purpose is not only to provide definitions of words, but also semantical relations between them, quasisynonymy, antonymy and hypernymy being the most important. 

From these 8 dictionaries, we built 8 AMR digraphs using the ideas described in Section \ref{sec:amr-digraphs}.
Each symbol-definition pairs were translated into AMR using the pre-trained model with the best \textit{smatch} \cite{cai2013smatch} made available through AMRlib \cite{jascob_amrlib_models_2023}.
The \textit{smatch} metric was proposed to evaluate the similarity between two AMR graphs by calculating the degree of overlap in their semantic structures.
It is commonly used to measure the precision of AMR StoG models: Jascob \cite{jascob_amrlib_models_2023} reports \textit{bart} as the best pre-trained StoG model and mentions - without data - that for the time being, encoder-decoder architecture outperforms readily available decoder-only models (models like Llama and ChatGPT) for the task of translating English sentences into AMR. 
Inferring the AMR digraphs from hundreds of thousands of English sentences was a non-trivial task and were thus computed on the Narval supercomputer, a high-performance computing system provided by Compute Canada. All manipulations of AMR digraphs were done through the Penman library \cite{goodman2020penman} and Networkx \cite{hagberg2008exploring}.

\begin{figure}[tb]
    \centering \includegraphics[width=0.5\textwidth]{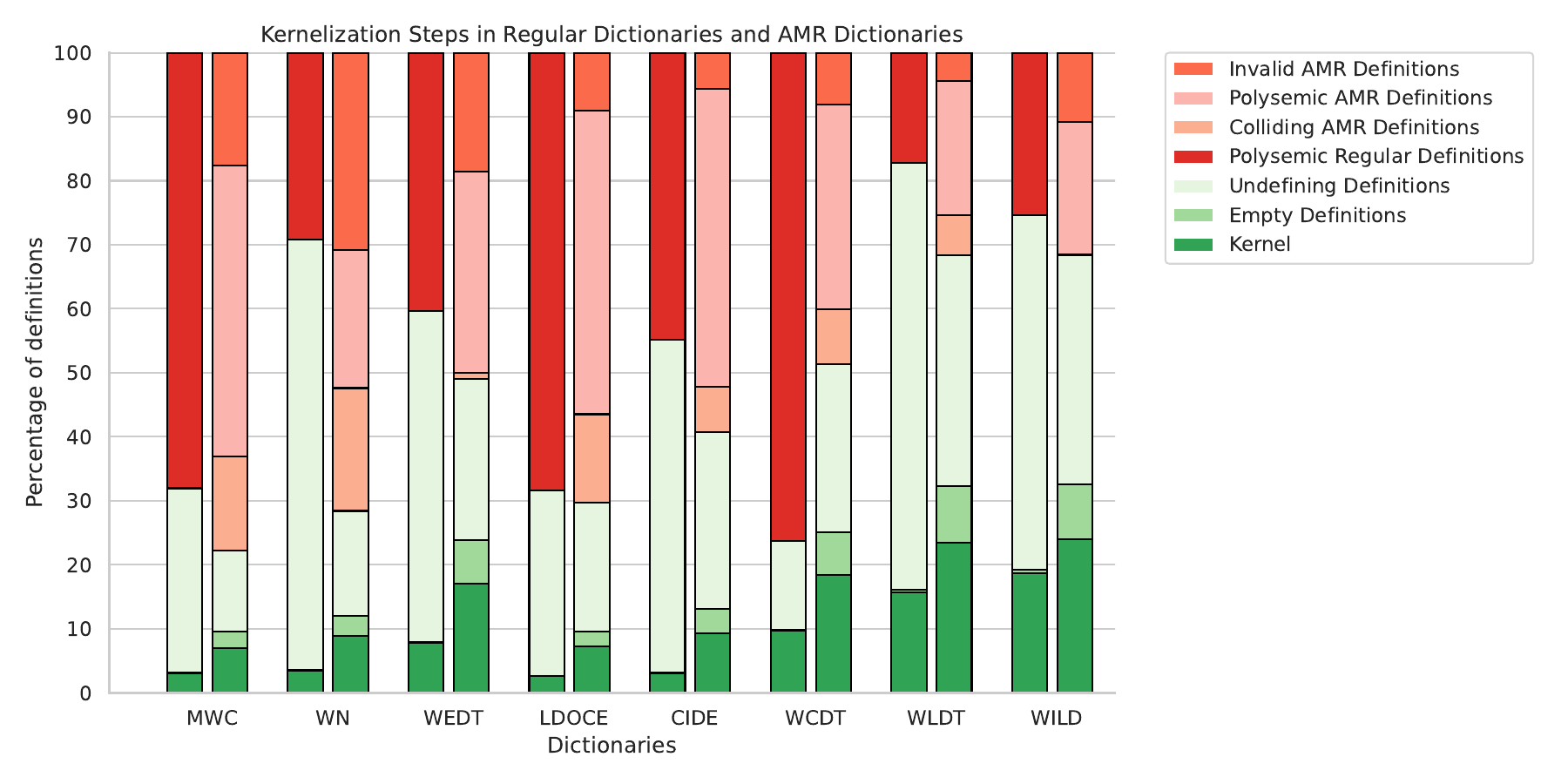}
    \caption{Shades of red in the upper part show number of definitions removed when starting with a complete digital dictionary and shades of green in the lower show the following preprocessing reducing it to its kernel. In each case, the left boxes show the values for the regular dictionaries, while the right boxes show the values for the AMR dictionaries.}
    \label{fig:dico-to-kernel}
\end{figure}

In order to compare our results with those described in \cite{vincent2016latent}, we also created digraphs from dictionaries using only the first definition of a given word.
Only content words, i.e. open-class words, were considered. 
Each definition was transformed into a graph where words are nodes and definitional relationships are arcs. The incoming arcs of a given node correspond to the words used in its definition and the outgoing arcs of a given node point to words it defines.
Figure \ref{fig:dico-to-kernel} details the effect of applying those transformations for each of the 8 dictionaries.

\begin{figure}[tb]
    \centering
    \includegraphics[width=0.5\textwidth]{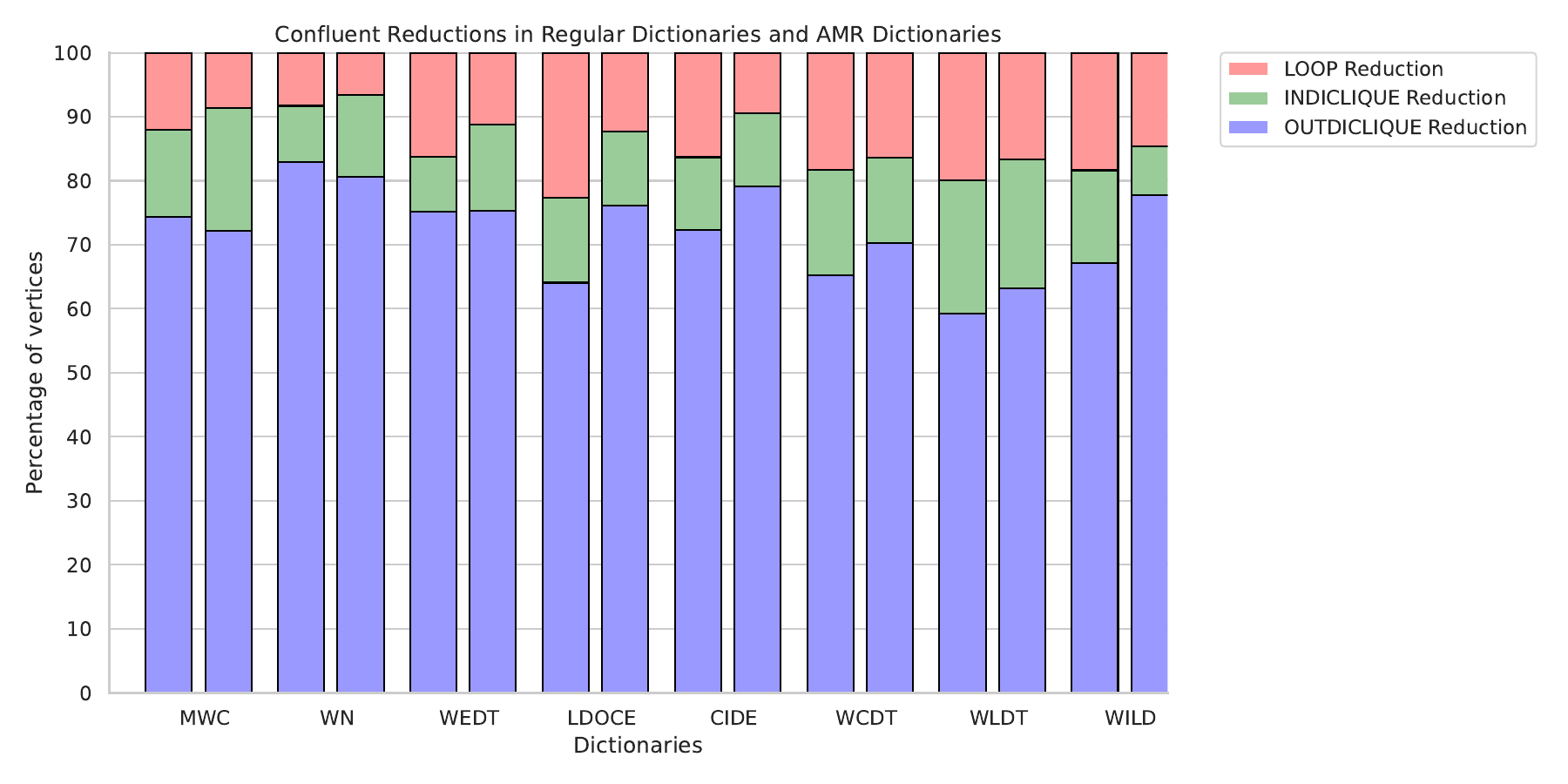}
    \caption{Proportion of applied confluent reductions in regular dictionaries (on the left) and AMR dictionaries (on the right).}
    \label{fig:reduction}
\end{figure}

Next, the reductions discussed in Section \ref{sec:reductions} were also applied on all 8 dictionaries.
In order to have a more universal reduction representative, we applied the confluent version ($A_c$) as a first step and then we applied the nonconfluent version ($A_{nc}$).
Figure \ref{fig:reduction} compares the proportion of application of each reduction, for each dictionary, with respect to regular dictionaries and AMR dictionaries, in the confluent case.
Note that we are only representing the predominant vertex reductions (excluding $\Subset$, as it is rarely applied). 
The focus is on reductions that affect the vertices of a dictionary or an AMR, because it is these latter which represent the \emph{words}. 
Since $\Pie$ and $\Dome$ are reductions that only impact edges, they are not represented in Figure~\ref{fig:reduction}.
Surprisingly, the majority of the reductions could be performed confluently.
Hence, we did not include the non confluent reductions in Figure \ref{fig:reduction}.
All algorithms related to reduction of directed graph were implemented in C++.
As both confluent and non confluent versions reduced all dictionaries in a matter of seconds or minutes, all computations were completed on a personal desktop.  

\begin{figure}[tb]
    \centering \includegraphics[width=0.5\textwidth]{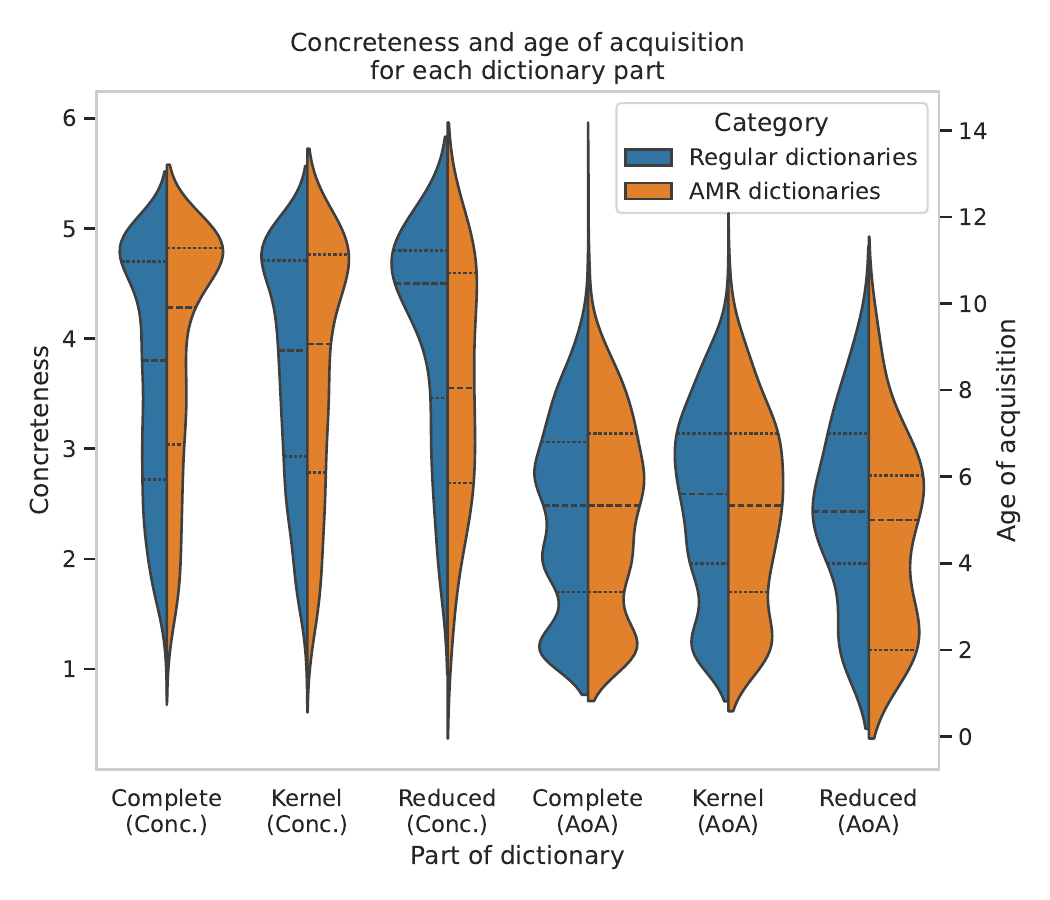}
    \caption{Distribution of the age of acquisition and concreteness for each type of dictionaries (regular and AMR) and each part of those dictionaries (complete, kernel and reduced).}
    \label{fig:psycho}
\end{figure}

Finally, we extracted all symbols or concepts common to all 8 digital dictionaries, for each dictionary part (complete, kernel and reduced) to study their psycholinguistical properties.
We restricted our attention to two key psycholinguistic variables: \emph{age of acquisition} (AOA) and \emph{concreteness} (Conc). 
AOA refers to the average age at which a word is first learned, with data sourced from Brysbaert's database (covering ages 1 to 21) \cite{brysbaert2017test}. 
Concreteness measures the materiality of a word, ranging from \emph{abstract} to \emph{highly concrete}, based on participants' classifications. Examples include "banana" and "apple" at level 5, while "belief" is rated as level 1 \cite{brysbaert2014concreteness}.
Figure \ref{fig:psycho} illustrates the evolution of the distribution of both the age of acquisition and the concreteness across all dictionary parts.


\section{Analysis and Discussion}\label{sec:discussion}

First, it is worth mentioning that while embedding into AMR causes less definitions to be included in the graph, at the end, they result in larger kernels. This suggests that AMR digraphs are more able to capture semantic and definitional relations via a less atomic symbolic representation of meaning.
Another notable observation is that the most applied reduction in all dictionaries, both AMR and regular, is $\Outdiclique$, as illustrated in Figure \ref{fig:reduction}.
This implies that many symbols or concepts used in some definitions can be replaced by others without losing the definitional path structure, the meaning of some implicit in others.
That fact is consistent with the idea behind the CIDE and LDOCE dictionaries, aiming to reduce the used words to a smaller \emph{controlled vocabulary}.

The violin plots illustrating the distribution of the age of acquisition and the concreteness yield interesting observations.
The effect of the reduction on regular dictionaries is not obvious, while a clearer tendency can be observed on AMR dictionary parts: As we reduce the directed graphs, the concepts are learned at a younger age and they become more abstract.
This is consistent with the intuition that words in the reduced digraph are more relevant to language learning.


\section{Concluding Remarks}

The main goal of this paper was to use the AMR formalism to study the symbol grounding problem.
The experimental results obtained seem to show that AMR is more or less well suited to improve WSD when studying this problem, but that it leads to more interpretable digraph structure thanks to its larger \textit{kernel}. Once minimal grounding sets are extracted from these, AMR arcs can be re-introduced to the remaining vertices, adding the notion of \textit{what relations between these words need to be known}. 
\newpage

\section{Limitations}

A first limitation lies in the imposed priority in Algorithm~\ref{algo:reduce-conf}. 
Although confluence guarantees a unique irreducible graph, the diagram in Figure~\ref{fig:reduction} and Table~\ref{tab:confl_reductions} might still differ because $\Indiclique$/$\Outdiclique$ and $\Pie$ for example could overlap. 
Specifically, a vertex with a single predecessor (or successor) that does not belong to any circuit can be identified by both $\Indiclique$ (or $\Outdiclique$) and $\Pie$, as all edges incident to this vertex are acyclic. 
From the perspective of studying the MFVS, it would be useful to exclude these vertices using $\Pie$ or $\Dome$ to assess how many vertices that appear in at least one circuit can be bypassed by $\Indiclique$/$\Outdiclique$.

To a lesser extent, another limitation concerns the interpretability of the reductions used from a psycholinguistic perspective. 
Only the simpler versions of the $\Indiclique$ and $\Outdiclique$ reductions can be interpreted when the vertex they apply to has at most one predecessor or successor. 
The removal of edges is mostly justified by the pursuit of the MFVS and lacks a clear psycholinguistic motivation.
A more in-depth study would have been to assess, among the various reductions, how often each one applies individually to the initial dictionary or AMR.

Even though confluence ensures uniqueness, the order in which reductions are applied needs further exploration, not only from the perspective of graph theory and MFVS but also considering psycholinguistics and the impact of these reductions on the structure of a dictionary and an AMR.

Another limitation is the proportion of rejected definitions. Whilst we set out to alleviate the issues caused by polysemy and could leverage AMR in novel ways, the proportion of preserved definitions was lesser in all dictionaries. The non-atomicity of AMR embeddings of the defined word in a definition proved to be the most common cause of an invalid AMR. Having non-atomic labels that can easily be identified when used to define another word makes it almost impossible to recreate the definitional relationships. This being said, the principal reason for rejecting definitions was the collision of labels.  Another issue lies in the limited number of symbols : AMR cannot contain all the possible definitions for all symbols included in the set of all dictionaries definitions. 

Next is the focus on English dictionaries. As we are writing this, we only had access to English dictionaries that could be analyzed. We are, however, very close to obtaining french dictionaries and we hope to extend our methods to dictionaries of different languages to see if dictionaries structures are similar across languages. We also hope to extend our semantic formalism to other languages by going from Abstract-Meaning-Representation to BabelNet-Meaning-Representation \cite{martinez-lorenzo-etal-2022-fully}, which is multilingual, multi-modal and disambiguated. 



\appendix 

\section{Directed Graphs}\label{section:notation}
This is the graph notation used to describe graph reductions in \ref{sec:reductions}.

A \emph{directed graph} or \emph{digraph} is an ordered pair $G = (V,A)$, where $V$ is a set whose elements are called \emph{vertices} and $E \subseteq A \times A$ is a set whose elements are called \emph{arcs}.
Another digraph $H = (V',A')$ is called a \emph{subgraph} of $G$ if $V' \subseteq V$ and $A' \subseteq \{(u,v) \in A \mid a,u \in V' \}$.
Let $G = (V, A)$, $u, v \in V$ and $U \subseteq V$.
We say that $u$ is a \emph{predecessor} (resp. \emph{successor}, resp. \emph{neighbor}) of $v$ if $(u, v) \in A$ (resp. $(v, u) \in A$, resp. $(u,v) \in A$ or $(v,u) \in A$).
The set of predecessors (resp. successors, resp. neighbors) of $u$ in $G$ is denoted by $N_G^-(u)$ (resp. $N_G^+(u)$, resp. $N_G(u)$).
Similarly, we say that the arc $a$ is an \emph{incoming} (resp. \emph{outgoing}, resp. \emph{incident}) arc of $u$ if $a = (u,v)$ (resp. $a = (v,u)$, resp. $a \in \{(u,v), (v,u)\}$) for some vertex $v$.
The set of incoming (resp. outgoing, resp. incident) arcs of $u$ in $G$ is denoted by $A_G^-(u)$ (resp. $A_G^+(u)$, resp. $A_G(u)$).
The set $U$ is called a \emph{directed clique} or \emph{diclique} of $G$ if $(u, v) \in A$ for all $u, v \in U$, $u \neq v$, and $(u, u) \notin U$, for all $u \in U$.
An arc $(u,v) \in A$ of $G$ is called \emph{bidirectional} if $(v,u) \in A$.
The set of all bidirectional arcs of $G$ is denoted by $\Abi$.

Given two vertices $u, v \in U$, a \emph{$uv$-path of $G$ of length $k$} is a sequence $p = (u_1, u_2, \ldots, u_{k-1})$, where $u = u_1$, $v = u_k$ and $(u_i, u_{i+1}) \in A$ for $i = 1, 2, \ldots, k - 2$.
We say that $p$ \emph{visits} a vertex $x$ (resp. an arc $(x,y)$) if $x = u_i$ for some $i \in \{1,2,\ldots,k\}$ (resp. $(x,y) = (u_i, u_{i+1})$ for some $i \in \{1,2,\ldots,k-2\}$).
If $u = v$, then $p$ is called a \emph{circuit}.
An arc is called \emph{acyclic} if there does not exist any circuit of $G$ that visits it.
Similarly, $G$ is called a \emph{directed acyclic graph} (DAG) if all its arcs are acyclic.
A digraph is called \emph{strongly connected} if for all $u,v \in V$ there exist at least one $uv$-path and at least one $vu$-path in $G$.
A \emph{strongly connected component} of a directed graph $G=(V,A)$ is a subgraph $H = (V',A')$ of $G$ that is strongly connected and such that $V'$ and $A'$ are maximal, meaning that no additional vertex or arc of $G$ can be included in $H$ without breaking its property of being strongly connected.

It is often convenient to consider digraphs obtained by removing some of its vertices or some of its arcs.
Let $G = (V,A)$ be a digraph, $u, v \in V$.
Then $G - u = (V', A')$ where $V' = V - \{u\}$ and $A' = A \cap V' \times V'$.
Similarly, $G - (u,v) = (V, A - \{(u,v)\})$.
The notation is extended to sets of vertices and sets of arcs: If $U \subseteq V$ and $B \subseteq A$, then $G - U = (V', A')$, where $V' = V - U$ and $A' = A \cap V' \times V'$ and $G - B = (V, A - B)$.
Finally, $G \circ u = (V', A')$, where $V' = V - u$ and $A' = (A - A_G(u)) \cup (N_G^-(u) \times N_G^+(u))$.

If $G=(V,A)$ is a digraph and $U$ a subset of $V$, we say that $U$ is a \emph{feedback vertex set} (\emph{FVS}) of $G$ if, for every circuit $c = (v_1, v_2, \ldots, v_k, v_1)$, there exists an index $i$ such that 
$v_i \in U$.
Equivalently, $U$ is a feedback vertex set of $G$ if $G - U$ is acyclic.
A \emph{minimum feedback vertex set} (\emph{MFVS}) of $G$ is a feedback vertex set of minimum cardinality. 
The problem of identifying MFVS in directed graphs is NP-hard and has been thoroughly studied in the last 40 years \cite{karp2010reducibility}.

\section{Results Tables}
This appendix section provides various results tables.


\begin{table*}
\centering
\resizebox{0.8\textwidth}{!}{%
\begin{tabular}{|l|ll|ll|ll|ll|ll|ll|ll|ll|}
\hline
           & \multicolumn{2}{c|}{WordNet}                                  & \multicolumn{2}{c|}{MerWeb}                                  & \multicolumn{2}{c|}{WEDT}                                    & \multicolumn{2}{c|}{LDOCE}                                   & \multicolumn{2}{c|}{CIDE}                                    & \multicolumn{2}{c|}{WCDT}                                   & \multicolumn{2}{c|}{WLDT}                                  & \multicolumn{2}{c|}{WILD}                                  \\ \hline
Nb Remain. & \multicolumn{1}{l|}{2 068}   & \cellcolor[HTML]{FFCCC9}2 948  & \multicolumn{1}{l|}{1 968}  & \cellcolor[HTML]{FFCCC9}2 846  & \multicolumn{1}{l|}{1 601}  & \cellcolor[HTML]{FFCCC9}1 836  & \multicolumn{1}{l|}{436}    & \cellcolor[HTML]{FFCCC9}931    & \multicolumn{1}{l|}{696}    & \cellcolor[HTML]{FFCCC9}1 067  & \multicolumn{1}{l|}{389}    & \cellcolor[HTML]{FFCCC9}374   & \multicolumn{1}{l|}{151}   & \cellcolor[HTML]{FFCCC9}95    & \multicolumn{1}{l|}{408}   & \cellcolor[HTML]{FFCCC9}216   \\ \hline
Nb Incl.   & \multicolumn{1}{l|}{548}     & \cellcolor[HTML]{FFCCC9}4 697  & \multicolumn{1}{l|}{1 125}  & \cellcolor[HTML]{FFCCC9}6 233  & \multicolumn{1}{l|}{899}    & \cellcolor[HTML]{FFCCC9}4 465  & \multicolumn{1}{l|}{390}    & \cellcolor[HTML]{FFCCC9}1 411  & \multicolumn{1}{l|}{185}    & \cellcolor[HTML]{FFCCC9}1 643  & \multicolumn{1}{l|}{460}    & \cellcolor[HTML]{FFCCC9}1 591 & \multicolumn{1}{l|}{208}   & \cellcolor[HTML]{FFCCC9}692   & \multicolumn{1}{l|}{172}   & \cellcolor[HTML]{FFCCC9}457   \\ \hline
Nb Excl.   & \multicolumn{1}{l|}{143 882} & \cellcolor[HTML]{FFCCC9}44 508 & \multicolumn{1}{l|}{94 402} & \cellcolor[HTML]{FFCCC9}50 332 & \multicolumn{1}{l|}{50 065} & \cellcolor[HTML]{FFCCC9}26 349 & \multicolumn{1}{l|}{34 773} & \cellcolor[HTML]{FFCCC9}19 527 & \multicolumn{1}{l|}{20 214} & \cellcolor[HTML]{FFCCC9}16 202 & \multicolumn{1}{l|}{11 530} & \cellcolor[HTML]{FFCCC9}8 655 & \multicolumn{1}{l|}{4 149} & \cellcolor[HTML]{FFCCC9}3 688 & \multicolumn{1}{l|}{2 633} & \cellcolor[HTML]{FFCCC9}2 391 \\ \hline
Nb Reduc.   & \multicolumn{1}{l|}{5 120}   & \cellcolor[HTML]{FFCCC9}3 969  & \multicolumn{1}{l|}{7 081}  & \cellcolor[HTML]{FFCCC9}4 879  & \multicolumn{1}{l|}{5 293}  & \cellcolor[HTML]{FFCCC9}3 386  & \multicolumn{1}{l|}{1 576}  & \cellcolor[HTML]{FFCCC9}1 291  & \multicolumn{1}{l|}{1 055}  & \cellcolor[HTML]{FFCCC9}998    & \multicolumn{1}{l|}{2 531}  & \cellcolor[HTML]{FFCCC9}1 545 & \multicolumn{1}{l|}{966}   & \cellcolor[HTML]{FFCCC9}812   & \multicolumn{1}{l|}{982}   & \cellcolor[HTML]{FFCCC9}463   \\ \hline
Nb LOOP    & \multicolumn{1}{l|}{409}     & \cellcolor[HTML]{FFCCC9}258    & \multicolumn{1}{l|}{818}    & \cellcolor[HTML]{FFCCC9}417    & \multicolumn{1}{l|}{814}    & \cellcolor[HTML]{FFCCC9}360    & \multicolumn{1}{l|}{341}    & \cellcolor[HTML]{FFCCC9}155    & \multicolumn{1}{l|}{162}    & \cellcolor[HTML]{FFCCC9}93     & \multicolumn{1}{l|}{415}    & \cellcolor[HTML]{FFCCC9}253   & \multicolumn{1}{l|}{185}   & \cellcolor[HTML]{FFCCC9}133   & \multicolumn{1}{l|}{160}   & \cellcolor[HTML]{FFCCC9}68    \\ \hline
Nb SUBSET  & \multicolumn{1}{l|}{3}       & \cellcolor[HTML]{FFCCC9}1      & \multicolumn{1}{l|}{2}      & \cellcolor[HTML]{FFCCC9}0      & \multicolumn{1}{l|}{3}      & \cellcolor[HTML]{FFCCC9}0      & \multicolumn{1}{l|}{3}      & \cellcolor[HTML]{FFCCC9}0      & \multicolumn{1}{l|}{2}      & \cellcolor[HTML]{FFCCC9}0      & \multicolumn{1}{l|}{4}      & \cellcolor[HTML]{FFCCC9}2     & \multicolumn{1}{l|}{1}     & \cellcolor[HTML]{FFCCC9}1     & \multicolumn{1}{l|}{1}     & \cellcolor[HTML]{FFCCC9}0     \\ \hline
Nb IN      & \multicolumn{1}{l|}{431}     & \cellcolor[HTML]{FFCCC9}495    & \multicolumn{1}{l|}{924}    & \cellcolor[HTML]{FFCCC9}929    & \multicolumn{1}{l|}{427}    & \cellcolor[HTML]{FFCCC9}433    & \multicolumn{1}{l|}{199}    & \cellcolor[HTML]{FFCCC9}147    & \multicolumn{1}{l|}{113}    & \cellcolor[HTML]{FFCCC9}113    & \multicolumn{1}{l|}{374}    & \cellcolor[HTML]{FFCCC9}205   & \multicolumn{1}{l|}{193}   & \cellcolor[HTML]{FFCCC9}160   & \multicolumn{1}{l|}{126}   & \cellcolor[HTML]{FFCCC9}35    \\ \hline
Nb OUT     & \multicolumn{1}{l|}{4 084}   & \cellcolor[HTML]{FFCCC9}3 137  & \multicolumn{1}{l|}{5 051}  & \cellcolor[HTML]{FFCCC9}3 487  & \multicolumn{1}{l|}{3 763}  & \cellcolor[HTML]{FFCCC9}2 414  & \multicolumn{1}{l|}{963}    & \cellcolor[HTML]{FFCCC9}961    & \multicolumn{1}{l|}{717}    & \cellcolor[HTML]{FFCCC9}779    & \multicolumn{1}{l|}{1 477}  & \cellcolor[HTML]{FFCCC9}1 082 & \multicolumn{1}{l|}{550}   & \cellcolor[HTML]{FFCCC9}504   & \multicolumn{1}{l|}{585}   & \cellcolor[HTML]{FFCCC9}360   \\ \hline
Nb PIE     & \multicolumn{1}{l|}{193}     & \cellcolor[HTML]{FFCCC9}78     & \multicolumn{1}{l|}{284}    & \cellcolor[HTML]{FFCCC9}46     & \multicolumn{1}{l|}{282}    & \cellcolor[HTML]{FFCCC9}179    & \multicolumn{1}{l|}{70}     & \cellcolor[HTML]{FFCCC9}28     & \multicolumn{1}{l|}{61}     & \cellcolor[HTML]{FFCCC9}13     & \multicolumn{1}{l|}{259}    & \cellcolor[HTML]{FFCCC9}3     & \multicolumn{1}{l|}{36}    & \cellcolor[HTML]{FFCCC9}14    & \multicolumn{1}{l|}{110}   & \cellcolor[HTML]{FFCCC9}0     \\ \hline
Nb ISOL.   & \multicolumn{1}{l|}{0}       & \cellcolor[HTML]{FFCCC9}0      & \multicolumn{1}{l|}{2}      & \cellcolor[HTML]{FFCCC9}0      & \multicolumn{1}{l|}{4}      & \cellcolor[HTML]{FFCCC9}0      & \multicolumn{1}{l|}{0}      & \cellcolor[HTML]{FFCCC9}0      & \multicolumn{1}{l|}{0}      & \cellcolor[HTML]{FFCCC9}0      & \multicolumn{1}{l|}{0}      & \cellcolor[HTML]{FFCCC9}0     & \multicolumn{1}{l|}{1}     & \cellcolor[HTML]{FFCCC9}0     & \multicolumn{1}{l|}{0}     & \cellcolor[HTML]{FFCCC9}0     \\ \hline
\end{tabular}
}
\caption{Confluent reductions metrics. Left column: regular dictionaries. Right column: AMR dictionaries.}\label{tab:confl_reductions}
\end{table*}

Table~\ref{tab:confl_reductions} presents the detailed results of the applications of different confluent reductions on dictionaries (white columns) and AMRs (red columns). Below are the details of each row:
\begin{itemize}
    \item Nb Remain. The number of vertices remaining in the irreducible graph.
    \item Nb Incl. The number of vertices included in the partial MFVS.
    \item Nb Excl. The number of excluded vertices.
    \item Nb Reduc. The total number of reductions applied.
    \item Nb $R$. The number of times the reduction $R$ was applied.
    \item Nb ISOL. The number of vertices whose incident edges have all been removed by $\Pie$ or $\Dome$.
\end{itemize}



\begin{table*}
\centering
\resizebox{0.8\textwidth}{!}{%
\begin{tabular}{|l|ll|ll|ll|ll|ll|ll|ll|ll|}
\hline
           & \multicolumn{2}{c|}{WordNet}                                  & \multicolumn{2}{c|}{MerWeb}                                  & \multicolumn{2}{c|}{WEDT}                                    & \multicolumn{2}{c|}{LDOCE}                                   & \multicolumn{2}{c|}{CIDE}                                    & \multicolumn{2}{c|}{WCDT}                                   & \multicolumn{2}{c|}{WLDT}                                  & \multicolumn{2}{c|}{WILD}                                  \\ \hline
Nb Remain. & \multicolumn{1}{l|}{2 028}   & \cellcolor[HTML]{FFCCC9}2 928  & \multicolumn{1}{l|}{1 933}  & \cellcolor[HTML]{FFCCC9}2 807  & \multicolumn{1}{l|}{1 556}  & \cellcolor[HTML]{FFCCC9}1 822  & \multicolumn{1}{l|}{423}    & \cellcolor[HTML]{FFCCC9}915    & \multicolumn{1}{l|}{683}    & \cellcolor[HTML]{FFCCC9}1 058  & \multicolumn{1}{l|}{342}    & \cellcolor[HTML]{FFCCC9}353   & \multicolumn{1}{l|}{127}   & \cellcolor[HTML]{FFCCC9}85    & \multicolumn{1}{l|}{398}   & \cellcolor[HTML]{FFCCC9}184   \\ \hline
Nb Incl.   & \multicolumn{1}{l|}{552}     & \cellcolor[HTML]{FFCCC9}4 697  & \multicolumn{1}{l|}{1 127}  & \cellcolor[HTML]{FFCCC9}6 234  & \multicolumn{1}{l|}{907}    & \cellcolor[HTML]{FFCCC9}4 466  & \multicolumn{1}{l|}{394}    & \cellcolor[HTML]{FFCCC9}1 412  & \multicolumn{1}{l|}{187}    & \cellcolor[HTML]{FFCCC9}1 644  & \multicolumn{1}{l|}{466}    & \cellcolor[HTML]{FFCCC9}1 594 & \multicolumn{1}{l|}{213}   & \cellcolor[HTML]{FFCCC9}693   & \multicolumn{1}{l|}{173}   & \cellcolor[HTML]{FFCCC9}466   \\ \hline
Nb Excl.   & \multicolumn{1}{l|}{143 918} & \cellcolor[HTML]{FFCCC9}44 528 & \multicolumn{1}{l|}{94 435} & \cellcolor[HTML]{FFCCC9}50 370 & \multicolumn{1}{l|}{50 102} & \cellcolor[HTML]{FFCCC9}26 362 & \multicolumn{1}{l|}{34 782} & \cellcolor[HTML]{FFCCC9}19 542 & \multicolumn{1}{l|}{20 225} & \cellcolor[HTML]{FFCCC9}16 210 & \multicolumn{1}{l|}{11 571} & \cellcolor[HTML]{FFCCC9}8 673 & \multicolumn{1}{l|}{4 168} & \cellcolor[HTML]{FFCCC9}3 697 & \multicolumn{1}{l|}{2 642} & \cellcolor[HTML]{FFCCC9}2 414 \\ \hline
Nb Reduc.   & \multicolumn{1}{l|}{5 498}   & \cellcolor[HTML]{FFCCC9}4 481  & \multicolumn{1}{l|}{7 526}  & \cellcolor[HTML]{FFCCC9}5 421  & \multicolumn{1}{l|}{5 733}  & \cellcolor[HTML]{FFCCC9}3 653  & \multicolumn{1}{l|}{1 727}  & \cellcolor[HTML]{FFCCC9}1 488  & \multicolumn{1}{l|}{1 184}  & \cellcolor[HTML]{FFCCC9}1 253  & \multicolumn{1}{l|}{2 828}  & \cellcolor[HTML]{FFCCC9}1 721 & \multicolumn{1}{l|}{1 086} & \cellcolor[HTML]{FFCCC9}879   & \multicolumn{1}{l|}{1 180} & \cellcolor[HTML]{FFCCC9}827   \\ \hline
Nb LOOP    & \multicolumn{1}{l|}{412}     & \cellcolor[HTML]{FFCCC9}258    & \multicolumn{1}{l|}{820}    & \cellcolor[HTML]{FFCCC9}418    & \multicolumn{1}{l|}{820}    & \cellcolor[HTML]{FFCCC9}361    & \multicolumn{1}{l|}{345}    & \cellcolor[HTML]{FFCCC9}156    & \multicolumn{1}{l|}{164}    & \cellcolor[HTML]{FFCCC9}93     & \multicolumn{1}{l|}{420}    & \cellcolor[HTML]{FFCCC9}256   & \multicolumn{1}{l|}{185}   & \cellcolor[HTML]{FFCCC9}134   & \multicolumn{1}{l|}{161}   & \cellcolor[HTML]{FFCCC9}77    \\ \hline
Nb SUBSET  & \multicolumn{1}{l|}{4}       & \cellcolor[HTML]{FFCCC9}1      & \multicolumn{1}{l|}{2}      & \cellcolor[HTML]{FFCCC9}0      & \multicolumn{1}{l|}{5}      & \cellcolor[HTML]{FFCCC9}0      & \multicolumn{1}{l|}{3}      & \cellcolor[HTML]{FFCCC9}0      & \multicolumn{1}{l|}{2}      & \cellcolor[HTML]{FFCCC9}1      & \multicolumn{1}{l|}{7}      & \cellcolor[HTML]{FFCCC9}2     & \multicolumn{1}{l|}{1}     & \cellcolor[HTML]{FFCCC9}1     & \multicolumn{1}{l|}{1}     & \cellcolor[HTML]{FFCCC9}0     \\ \hline
Nb IN      & \multicolumn{1}{l|}{457}     & \cellcolor[HTML]{FFCCC9}504    & \multicolumn{1}{l|}{940}    & \cellcolor[HTML]{FFCCC9}945    & \multicolumn{1}{l|}{441}    & \cellcolor[HTML]{FFCCC9}442    & \multicolumn{1}{l|}{205}    & \cellcolor[HTML]{FFCCC9}155    & \multicolumn{1}{l|}{118}    & \cellcolor[HTML]{FFCCC9}117    & \multicolumn{1}{l|}{394}    & \cellcolor[HTML]{FFCCC9}185   & \multicolumn{1}{l|}{206}   & \cellcolor[HTML]{FFCCC9}111   & \multicolumn{1}{l|}{133}   & \cellcolor[HTML]{FFCCC9}53    \\ \hline
Nb OUT     & \multicolumn{1}{l|}{4 093}   & \cellcolor[HTML]{FFCCC9}3 146  & \multicolumn{1}{l|}{5 067}  & \cellcolor[HTML]{FFCCC9}3 509  & \multicolumn{1}{l|}{3 789}  & \cellcolor[HTML]{FFCCC9}2 418  & \multicolumn{1}{l|}{966}    & \cellcolor[HTML]{FFCCC9}968    & \multicolumn{1}{l|}{723}    & \cellcolor[HTML]{FFCCC9}783    & \multicolumn{1}{l|}{1 497}  & \cellcolor[HTML]{FFCCC9}1 085 & \multicolumn{1}{l|}{556}   & \cellcolor[HTML]{FFCCC9}607   & \multicolumn{1}{l|}{587}   & \cellcolor[HTML]{FFCCC9}364   \\ \hline
Nb PIE     & \multicolumn{1}{l|}{193}     & \cellcolor[HTML]{FFCCC9}78     & \multicolumn{1}{l|}{284}    & \cellcolor[HTML]{FFCCC9}46     & \multicolumn{1}{l|}{298}    & \cellcolor[HTML]{FFCCC9}179    & \multicolumn{1}{l|}{70}     & \cellcolor[HTML]{FFCCC9}28     & \multicolumn{1}{l|}{61}     & \cellcolor[HTML]{FFCCC9}13     & \multicolumn{1}{l|}{259}    & \cellcolor[HTML]{FFCCC9}3     & \multicolumn{1}{l|}{36}    & \cellcolor[HTML]{FFCCC9}60    & \multicolumn{1}{l|}{110}   & \cellcolor[HTML]{FFCCC9}4     \\ \hline
Nb DOME++  & \multicolumn{1}{l|}{338}     & \cellcolor[HTML]{FFCCC9}492    & \multicolumn{1}{l|}{410}    & \cellcolor[HTML]{FFCCC9}503    & \multicolumn{1}{l|}{379}    & \cellcolor[HTML]{FFCCC9}253    & \multicolumn{1}{l|}{138}    & \cellcolor[HTML]{FFCCC9}181    & \multicolumn{1}{l|}{116}    & \cellcolor[HTML]{FFCCC9}246    & \multicolumn{1}{l|}{250}    & \cellcolor[HTML]{FFCCC9}155   & \multicolumn{1}{l|}{96}    & \cellcolor[HTML]{FFCCC9}165   & \multicolumn{1}{l|}{188}   & \cellcolor[HTML]{FFCCC9}328   \\ \hline
Nb ISOL.   & \multicolumn{1}{l|}{1}       & \cellcolor[HTML]{FFCCC9}2      & \multicolumn{1}{l|}{3}      & \cellcolor[HTML]{FFCCC9}0      & \multicolumn{1}{l|}{1}      & \cellcolor[HTML]{FFCCC9}0      & \multicolumn{1}{l|}{0}      & \cellcolor[HTML]{FFCCC9}0      & \multicolumn{1}{l|}{0}      & \cellcolor[HTML]{FFCCC9}0      & \multicolumn{1}{l|}{1}      & \cellcolor[HTML]{FFCCC9}0     & \multicolumn{1}{l|}{1}     & \cellcolor[HTML]{FFCCC9}0     & \multicolumn{1}{l|}{0}     & \cellcolor[HTML]{FFCCC9}1     \\ \hline
\end{tabular}
}
\caption{Non-confluent reductions metrics. Left column: regular dictionaries. Right column: AMR dictionaries.}\label{tab:nc_reductions}
\end{table*}

Like Table~\ref{tab:confl_reductions}, Table~\ref{tab:nc_reductions} presents the detailed results of the applications of different non-confluent reductions on dictionaries (white columns) and AMRs (red columns).
The rows have the same meaning as in Table~\ref{tab:confl_reductions}.


\begin{table*}
\centering
\resizebox{0.8\textwidth}{!}{%
\begin{tabular}{|l|ll|ll|ll|ll|ll|ll|ll|ll|}
\hline
               & \multicolumn{2}{c|}{WordNet}                                  & \multicolumn{2}{c|}{MerWeb}                                    & \multicolumn{2}{c|}{WEDT}                                      & \multicolumn{2}{c|}{LDOCE}                                     & \multicolumn{2}{c|}{CIDE}                                      & \multicolumn{2}{c|}{WCDT}                                    & \multicolumn{2}{c|}{WLDT}                                    & \multicolumn{2}{c|}{WILD}                                     \\ \hline
Nb vertices    & \multicolumn{1}{l|}{146 498} & \cellcolor[HTML]{FFCCC9}52 153 & \multicolumn{1}{l|}{97 495}  & \cellcolor[HTML]{FFCCC9}59 411  & \multicolumn{1}{l|}{52 565}  & \cellcolor[HTML]{FFCCC9}32 650  & \multicolumn{1}{l|}{35 599}  & \cellcolor[HTML]{FFCCC9}21 869  & \multicolumn{1}{l|}{21 095}  & \cellcolor[HTML]{FFCCC9}18 912  & \multicolumn{1}{l|}{12 379} & \cellcolor[HTML]{FFCCC9}10 620 & \multicolumn{1}{l|}{4 508}  & \cellcolor[HTML]{FFCCC9}4 475  & \multicolumn{1}{l|}{3 213}  & \cellcolor[HTML]{FFCCC9}3 064   \\ \hline
Size Kernel    & \multicolumn{1}{l|}{7 131}   & \cellcolor[HTML]{FFCCC9}11 277 & \multicolumn{1}{l|}{9 070}   & \cellcolor[HTML]{FFCCC9}13 495  & \multicolumn{1}{l|}{6 694}   & \cellcolor[HTML]{FFCCC9}9 148   & \multicolumn{1}{l|}{1 988}   & \cellcolor[HTML]{FFCCC9}3 450   & \multicolumn{1}{l|}{1 711}   & \cellcolor[HTML]{FFCCC9}3 602   & \multicolumn{1}{l|}{2 813}  & \cellcolor[HTML]{FFCCC9}3 252  & \multicolumn{1}{l|}{1 103}  & \cellcolor[HTML]{FFCCC9}1 451  & \multicolumn{1}{l|}{1 291}  & \cellcolor[HTML]{FFCCC9}1 068   \\ \hline
Size Red. Ker. & \multicolumn{1}{l|}{2 068}   & \cellcolor[HTML]{FFCCC9}2 948  & \multicolumn{1}{l|}{1 968}   & \cellcolor[HTML]{FFCCC9}2 846   & \multicolumn{1}{l|}{1 601}   & \cellcolor[HTML]{FFCCC9}1 836   & \multicolumn{1}{l|}{436}     & \cellcolor[HTML]{FFCCC9}931     & \multicolumn{1}{l|}{696}     & \cellcolor[HTML]{FFCCC9}1 067   & \multicolumn{1}{l|}{389}    & \cellcolor[HTML]{FFCCC9}374    & \multicolumn{1}{l|}{147}    & \cellcolor[HTML]{FFCCC9}95     & \multicolumn{1}{l|}{408}    & \cellcolor[HTML]{FFCCC9}216     \\ \hline
NC Red Ker.    & \multicolumn{1}{l|}{2 028}   & \cellcolor[HTML]{FFCCC9}2 928  & \multicolumn{1}{l|}{1 933}   & \cellcolor[HTML]{FFCCC9}2 807   & \multicolumn{1}{l|}{1 556}   & \cellcolor[HTML]{FFCCC9}1 822   & \multicolumn{1}{l|}{423}     & \cellcolor[HTML]{FFCCC9}915     & \multicolumn{1}{l|}{683}     & \cellcolor[HTML]{FFCCC9}1 058   & \multicolumn{1}{l|}{342}    & \cellcolor[HTML]{FFCCC9}353    & \multicolumn{1}{l|}{123}    & \cellcolor[HTML]{FFCCC9}85     & \multicolumn{1}{l|}{398}    & \cellcolor[HTML]{FFCCC9}184     \\ \hline
Init. Nb Arcs  & \multicolumn{1}{l|}{765 481} & \cellcolor[HTML]{FFCCC9}311321 & \multicolumn{1}{l|}{495 385} & \cellcolor[HTML]{FFCCC9}337 376 & \multicolumn{1}{l|}{273 432} & \cellcolor[HTML]{FFCCC9}190 470 & \multicolumn{1}{l|}{173 737} & \cellcolor[HTML]{FFCCC9}147 319 & \multicolumn{1}{l|}{107 638} & \cellcolor[HTML]{FFCCC9}144 308 & \multicolumn{1}{l|}{61 282} & \cellcolor[HTML]{FFCCC9}56 102 & \multicolumn{1}{l|}{19 133} & \cellcolor[HTML]{FFCCC9}21 683 & \multicolumn{1}{l|}{22 667} & \cellcolor[HTML]{FFCCC9}16 579  \\ \hline
Final Nb Arcs  & \multicolumn{1}{l|}{14 858}  & \cellcolor[HTML]{FFCCC9}32 277 & \multicolumn{1}{l|}{14 278}  & \cellcolor[HTML]{FFCCC9}33 268  & \multicolumn{1}{l|}{11 109}  & \cellcolor[HTML]{FFCCC9}19 147  & \multicolumn{1}{l|}{2 700}   & \cellcolor[HTML]{FFCCC9}8 943   & \multicolumn{1}{l|}{4 837}   & \cellcolor[HTML]{FFCCC9}11 940  & \multicolumn{1}{l|}{2 524}  & \cellcolor[HTML]{FFCCC9}3 665  & \multicolumn{1}{l|}{833}    & \cellcolor[HTML]{FFCCC9}1 149  & \multicolumn{1}{l|}{2 860}  & \cellcolor[HTML]{FFCCC9}1 828   \\ \hline
NC Final Arcs  & \multicolumn{1}{l|}{14 430}  & \cellcolor[HTML]{FFCCC9}31 766 & \multicolumn{1}{l|}{13 831}  & \cellcolor[HTML]{FFCCC9}32 754  & \multicolumn{1}{l|}{10 520}  & \cellcolor[HTML]{FFCCC9}18 849  & \multicolumn{1}{l|}{2 539}   & \cellcolor[HTML]{FFCCC9}8 741   & \multicolumn{1}{l|}{4 697}   & \cellcolor[HTML]{FFCCC9}11 660  & \multicolumn{1}{l|}{2 177}  & \cellcolor[HTML]{FFCCC9}3 460  & \multicolumn{1}{l|}{703}    & \cellcolor[HTML]{FFCCC9}1 076  & \multicolumn{1}{l|}{2 664}  & \cellcolor[HTML]{FFCCC9}1 424   \\ \hline
Nb Undefined   & \multicolumn{1}{l|}{144}     & \cellcolor[HTML]{FFCCC9}4 504  & \multicolumn{1}{l|}{328}     & \cellcolor[HTML]{FFCCC9}6 010   & \multicolumn{1}{l|}{89}      & \cellcolor[HTML]{FFCCC9}4 309   & \multicolumn{1}{l|}{46}      & \cellcolor[HTML]{FFCCC9}1 279   & \multicolumn{1}{l|}{22}      & \cellcolor[HTML]{FFCCC9}1 565   & \multicolumn{1}{l|}{48}     & \cellcolor[HTML]{FFCCC9}1 365  & \multicolumn{1}{l|}{33}     & \cellcolor[HTML]{FFCCC9}573    & \multicolumn{1}{l|}{11}     & \cellcolor[HTML]{FFCCC9}389     \\ \hline
Nb Undefining  & \multicolumn{1}{l|}{137 815} & \cellcolor[HTML]{FFCCC9}40 509 & \multicolumn{1}{l|}{86 374}  & \cellcolor[HTML]{FFCCC9}44 626  & \multicolumn{1}{l|}{44 280}  & \cellcolor[HTML]{FFCCC9}22 863  & \multicolumn{1}{l|}{33 246}  & \cellcolor[HTML]{FFCCC9}18 261  & \multicolumn{1}{l|}{18 843}  & \cellcolor[HTML]{FFCCC9}15 105  & \multicolumn{1}{l|}{9 278}  & \cellcolor[HTML]{FFCCC9}7 316  & \multicolumn{1}{l|}{3 261}  & \cellcolor[HTML]{FFCCC9}2997   & \multicolumn{1}{l|}{1 874}  & \cellcolor[HTML]{FFCCC9}1 991   \\ \hline
Nb SCCs Kernel & \multicolumn{1}{l|}{253}     & \cellcolor[HTML]{FFCCC9}81     & \multicolumn{1}{l|}{459}     & \cellcolor[HTML]{FFCCC9}188     & \multicolumn{1}{l|}{385}     & \cellcolor[HTML]{FFCCC9}112     & \multicolumn{1}{l|}{168}     & \cellcolor[HTML]{FFCCC9}71      & \multicolumn{1}{l|}{65}      & \cellcolor[HTML]{FFCCC9}23      & \multicolumn{1}{l|}{125}    & \cellcolor[HTML]{FFCCC9}68     & \multicolumn{1}{l|}{49}     & \cellcolor[HTML]{FFCCC9}27     & \multicolumn{1}{l|}{18}     & \cellcolor[HTML]{FFCCC9}22      \\ \hline
Kernel Nb Arcs & \multicolumn{1}{l|}{33 284}  & \cellcolor[HTML]{FFCCC9}50 620 & \multicolumn{1}{l|}{46 405}  & \cellcolor[HTML]{FFCCC9}58 260  & \multicolumn{1}{l|}{33 580}  & \cellcolor[HTML]{FFCCC9}35 308  & \multicolumn{1}{l|}{10 014}  & \cellcolor[HTML]{FFCCC9}16 058  & \multicolumn{1}{l|}{9 792}   & \cellcolor[HTML]{FFCCC9}17 459  & \multicolumn{1}{l|}{13 929} & \cellcolor[HTML]{FFCCC9}10 295 & \multicolumn{1}{l|}{4 459}  & \cellcolor[HTML]{FFCCC9}4 306  & \multicolumn{1}{l|}{8 618}  & \cellcolor[HTML]{FFCCC9}3 728   \\ \hline
Kernel Density & \multicolumn{1}{l|}{0.0007}  & \cellcolor[HTML]{FFCCC9}0.0004 & \multicolumn{1}{l|}{0.0006}  & \cellcolor[HTML]{FFCCC9}0.0003  & \multicolumn{1}{l|}{0.0007}  & \cellcolor[HTML]{FFCCC9}0.0004  & \multicolumn{1}{l|}{0.0025}  & \cellcolor[HTML]{FFCCC9}0.0014  & \multicolumn{1}{l|}{0.0034}  & \cellcolor[HTML]{FFCCC9}0.0014  & \multicolumn{1}{l|}{0.0018} & \cellcolor[HTML]{FFCCC9}0.0010 & \multicolumn{1}{l|}{0.0037} & \cellcolor[HTML]{FFCCC9}0.0020 & \multicolumn{1}{l|}{0.0052} & \cellcolor[HTML]{FFCCC9}0.0033 \\ \hline
\end{tabular}
}
\caption{Graph metrics. Left column: regular dictionaries. Right column: AMR dictionaries.}
\label{tab:graph_metrics}
\end{table*}

Table \ref{tab:graph_metrics} summarizes the metrics of the graphs created from the union of graphs made from individual definitions. Once again, regular dictionaries metrics are shown in white columns and AMR's in red columns. While it is true that less definitions are taken into account by AMR, it still has a somewhat more dense representation of meaning as even when taking into account the loss in definitions. It is also noteworthy that AMR embedding have a much smaller proportions of undefining words, implying what words appear are generally more often used. It also has a far higher proportion of undefined words : this is due to the fact that AMR is like translating from English to another dictionary. Many used symbols are left undefined as they are lifted from the Propbank Frames. This suggests a potential solution would be to source missing definitions in Propbank Frames. The kernels of regular dictionary are also generally almost twice as dense as those of AMR dictionaries. Here are the details of each row : 

\begin{itemize}
    \item Nb Vertices : The total number of vertices in the graph before reductions
    \item Size Kernel The number of vertices remaining once all non-defined and non-defining words have been recursively removes
    \item Size Red. Kernel : The number of vertices in the reduced kernel and is equivalent to the remaining vertices in the confluent reductions table.
    \item NC Red. Kernel : The number of vertices in the non-confluently reduced kernel. Is equivalent to the remaining vertices in the non-confluent reductions tables. 
    \item Init Nb Arcs : The initial number of arcs in the graph before reductions
    \item Final Nb Arcs : The number of arcs after confluent reductions
    \item NC Final Arcs : The number of arcs after non-confluent reductions
    \item Nb Undefined : The number of words removed when recursively removing words not defined by any other
    \item Nb Undefining : The number of words removed when recursively removing all words not used to define another. 
    \item Nb SCCs Kernel : The number of strongly connected components found in the kernel. All kernels are made up almost entirely of their biggest strongly connected component and all other SCCs are of negligible size. 
    \item Kernel Nb Arcs : The number of arcs for a given kernel 
    \item Kernel Density : The density of each kernel.
\end{itemize}


\begin{table*}
\centering
\resizebox{0.8\textwidth}{!}{%
\begin{tabular}{|l|l|l|l|l|l|l|l|l|}
\hline
                         & MerWeb  & WordNet & WEDT   & LDOCE  & CIDE   & WCDT   & WLDT  & WILD  \\ \hline
Definition Quantity      & 301 240 & 206 185 & 86 949 & 80 086 & 49 787 & 22 563 & 6 900 & 4 709 \\ \hline
Initial Invalid Quantity & 98 349  & 98 443  & 28 548 & 17 030 & 10 484 & 3 910  & 739   & 796   \\ \hline
Saved Quantity           & 43 632  & 32 195  & 13 721 & 9 538  & 7 564  & 1 936  & 400   & 231   \\ \hline
Final Invalid Quantity   & 54 717  & 66 248  & 14 827 & 7 492  & 2 920  & 1 974  & 339   & 565   \\ \hline
Polysemy Filtered        & 141 255 & 46 486  & 25 137 & 39 257 & 24 296 & 7 822  & 1 600 & 1 085 \\ \hline
Symbols Collisions       & 45 587  & 41 298  & 14 335 & 11 468 & 3 659  & 2 097  & 486   & 5     \\ \hline
Final Quantity           & 59 411  & 52 153  & 32 650 & 21 869 & 18 912 & 10 620 & 4 475 & 3 064 \\ \hline
\end{tabular}
}
\caption{AMR digraphs preprocessing metrics.}
\label{tab:graph_production}
\end{table*}

Table \ref{tab:graph_production} describes the metrics of creating AMR graphs from definitions. It is noteworthy that the principal reason for discarding a definition embedded in AMR is label collision, within a symbol's definitions and across symbols definitions. Here are the details for each row : 

\begin{itemize}
    \item Definition Quantity : The total number of definitions in a dictionary, including polysemic definitions (when a word has multiple definitions).
    \item Initial Invalid Quantity : The initial quantity of AMR graphs that were not valid as per our requirements described in \ref{sec:amr-digraphs}
    \item Save Quantity : The number of AMR graphs saved by making them \textit{valid} via our two strategies described in \ref{ss:format}.
    \item Final Invalid Quantity : The number of graphs that could not be made valid. 
    \item Polysemy Filtered : The number of polysemic definitions lost due to having the same label for the defined word. 
    \item Symbols Collisions : The number of definitions lost due to having the same label for the label for the defined word across different words.
    \item Final Quantity : The final quantity of definitions used in the graph.
\end{itemize}


\begin{table}[t]
  \scriptsize
  \centering
  \begin{tabular}{llllllllll}
  animal & fight & high & paper & thread \\
  attention & finger & law & picture & time \\
  bird & flat & leg & plant & touch \\
  building & food & letter & power & train \\
  control & foot & life & protect & tree \\
  curve & game & liquid & room & verb \\
  decide & gas & machine & season & water \\
  earth & glass & metal & sleep & wheel \\
  eat & grain & mind & soft & wing \\
  end & hand & money & solid & word \\
  face & happen & new & sun & work \\
  farm & head & number & taste & write \\
  female
  \end{tabular}
  \caption{List of regular dictionary symbols common to the 8 digital reduced dictionaries}\label{tab:common-dico}
\end{table}

\begin{table}[t]
  \scriptsize
  \centering
  \begin{tabular}{llllllll}
  another & have-degree-91 & position \\
  area & have-frequency-91 & product-of \\
  ask-02 & have-org-role-91 & real-04 \\
  at-a-time & have-part-91 & regular-03 \\
  be-located-at-91 & have-quant-91 & relative-position \\
  brown-01 & have-rel-role-91 & responsible-03 \\
  brown-02 & information & return-05 \\
  building & in-front-of & right \\
  careful & join-01 & road \\
  catch-01 & keep-04 & round-06 \\
  clothes & know-02 & rule \\
  come-03 & leg & same-01 \\
  complete-02 & make-05 & send-03 \\
  concern-02 & marry-01 & size-01 \\
  copy-01 & mass-quantity & smooth-06 \\
  country & member & son \\
  cover-02 & mind & space \\
  cut-01 & most & speak-01 \\
  danger & mouth & spend-02 \\
  dark-03 & new-01 & stiff-04 \\
  date-entity & number & string-entity \\
  date-interval & officer & sweet-04 \\
  decide-01 & one & temporal-quantity \\
  distance-quantity & only & that \\
  dry-08 & open-06 & thing \\
  energy & open-09 & together \\
  enough & opening & top \\
  equal-01 & ordinal-entity & touch-01 \\
  every & part & use-03 \\
  feel-01 & particular & value-02 \\
  fit-06 & person & water \\
  food & picture & white-03 \\
  force-01 & piece & whole \\
  force-04 & place-01 & word \\
  get-04 & planet & work-09 \\
  good-03 & plant & write-01 \\
  grass & play-02 & wrong-02 \\
  group & play-11 & yellow-02 \\
  happy-02
  \end{tabular}
  \caption{List of AMR concepts common to the 8 digital reduced dictionaries}\label{tab:common-amr}
\end{table}

Table \ref{tab:common-dico} gives the list of the symbols common to the 8 reduced regular dictionaries.
Similarly, Table \ref{tab:common-amr} gives the AMR concepts occurring in each of the 8 reduced AMR dictionaries.

\end{document}